\documentclass[journal]{IEEEtran}
\usepackage{graphicx}
\graphicspath{{figures/}} 
\usepackage[export]{adjustbox}
\usepackage{times}
\usepackage{epsfig}
\usepackage{graphicx}
\usepackage{amsmath}
\usepackage{amssymb}
\usepackage{color}
\usepackage{caption}
\usepackage{subcaption}
\usepackage{float}

\definecolor{darkgreen}{rgb}{0,0.6,0.2}

\usepackage[pagebackref=true,breaklinks=true,letterpaper=true,colorlinks,bookmarks=false]{hyperref}

\begin{document}


\title{




\textit{This} looks \textit{more} like \textit{that}: Enhancing Self-Explaining Models by Prototypical Relevance Propagation 
}


\author{{Srishti~Gautam,
        Marina~M.-C. Höhne,
        Stine~Hansen,
        Robert~Jenssen and
        Michael~Kampffmeyer}
\thanks{This work was financially supported by the Research Council of Norway (RCN), through its Centre for Research-based Innovation funding scheme (Visual Intelligence, grant no. 309439), and Consortium Partners. The work was further partially funded by RCN FRIPRO grant no. 315029 and RCN IKTPLUSS grant no. 303514.}%
\thanks{Srishti Gautam, Stine Hansen, Robert Jenssen and Michael Kampffmeyer are with the Department of Physics and Technology, UiT The Arctic University of Norway, 9037 Tromsø, Norway (e-mail: srishti.gautam@uit.no; s.hansen@uit.no; robert.jenssen@uit.no;  michael.c.kampffmeyer@uit.no).}
\thanks{Marina~M.-C. Höhne is with the Institute of Software Engineering and Theoretical Computer Science at Technische Universität Berlin, 10623 Berlin, Germany (email: marina.hoehne@tu-berlin.de).}
}

\maketitle

\begin{abstract}
Current machine learning models have shown high efficiency in solving a wide variety of real-world problems. However, their black box character poses a major challenge for the understanding and traceability of the underlying decision-making strategies. As a remedy, many post-hoc explanation and self-explanatory methods have been developed to interpret the models' behavior. These methods, in addition, enable the identification of artifacts that can be learned by the model as class-relevant features.
In this work, we provide a detailed case study of the self-explaining network, ProtoPNet, in the presence of a spectrum of artifacts. Accordingly, we identify the main drawbacks of ProtoPNet, especially, its coarse and spatially imprecise explanations. We address these limitations by introducing Prototypical Relevance Propagation (PRP), a novel method for generating more precise model-aware explanations.
Furthermore, in order to obtain a clean dataset, we propose to use multi-view clustering strategies for segregating the artifact images using the PRP explanations, thereby suppressing the potential artifact learning in the models. 
\end{abstract}

\section{Introduction}
When applying AI models, especially in safety-critical areas, such as medical applications, autonomous driving, or criminal justice, we need to understand their underlying behavior to decide whether the models can be trusted or not. Here, the field of explainable AI (XAI) has established itself, where methods are being developed to illuminate the so-called black box models. XAI will serve as an essential support in ethical, legal, and social issues and ultimately also contribute to an increased acceptance by the end user. 

Recent efforts in the field of XAI have revealed undesirable behavior of AI models \cite{chans}. This especially includes, but is not limited to, a behavior called Clever Hans. A Clever Hans is a class specific artifact present in the training data and thus learned by the model as a relevant feature. 
Explanation methods, such as Layerwise Relevance Propagation (LRP) \cite{lrp} can highlight the input features that led the model to a particular behavior. These methods can therefore reveal those Clever Hans artifacts, which can then be visually identified by the user.
For example, the authors in \cite{chans} showed by the use of LRP that, for their specific dataset and model, none of the features of the horse were responsible for the prediction of the class ``horse". Rather, the features of the photographer's watermark were shown as relevant. This can be seen as ``cheating by the model" to achieve a lower loss and therefore a higher accuracy. 
Furthermore, problems such as adversarial or malicious attacks can arise, where an adversary applies perturbations to the data to force the model to deliver incorrect results. One of them is known as a backdoor attack \cite{backdoor}, in which the black box models react extremely unexpectedly from an unaware user's perspective but, as expected, maliciously from the attacker's perspective in the presence of backdoor triggers. This emphasizes the importance of finding and suppressing these artifacts either from the model's learnt representations or from the data itself. XAI can be one very effective method in discovering and suppressing these undesirable behaviors \cite{chans, imagenetCH}.

While the treatment of artifacts has received limited attention, \cite{chans, imagenetCH} propose the use of spectral relevance analysis (SpRAy) on the LRP explanation maps of data to detect strategies and suppress artifacts learned by the models.
The aforementioned LRP has received much attention and is a prominent example of the so called post-hoc XAI methods, which are methods developed separately from the black box models to explain their decisions.

However, recently, \cite{rudin2019stop} suggested to develop self-explaining neural networks. These intrinsically explain the decision making process to the user without a further need of post-hoc explanation methods. Towards this goal, the author behind \cite{rudin2019stop} proposed a network (ProtoPNet) that provides a transparent prediction by introducing a prototype layer between the final convolution layer and the output layer \cite{ppnet}. This prototype layer consists of a fixed number of prototypes for each class, which can be thought of as representative instances for each class of the training data. During the classification process, for each image that is passed through the network, an activation map based on its similarity with respect to every prototype is computed which is then used for the final classification. Afterwards, by upsampling the activation map to the input size, the 
most relevant pixels, contributing to the classification, are highlighted.
Doing this for both, the prototype (training) images and the test image, the regions of interests can be visualized which serve as a direct comparison for the user to capture the relation between the test image and the prototype images from the training set. This accordingly helps in comprehending the decision of the network by ``this relevant feature of the test image looks like that relevant feature from the class-specific prototype image".

Recalling the artifacts issue, the solution now appears to be clear when using self-explaining neural networks, such as ProtoPNet: If the model learned a feature corresponding to the artifact, then it must be reflected by at least one of the prototypes of the class consisting of such artifacts. After identifying these artifact prototypes, they can be pruned from the model thereby making the model artifact-free.

However, in this work we show that this is not a feasible approach because of coarse and spatially imprecise explanations provided by ProtoPNet due to its model-agnostic upsampling. 
Therefore, building on the principles of LRP, we propose a novel method referred to as Prototypical Relevance Propagation (PRP) to attain more accurate model-aware explanations. Using PRP, we illustrate that artifact information is entangled within the ProtoPNet, such that most prototypes capture artifact related features, making the above-mentioned pruning procedure unreasonable. 
Further, to preserve the strength of ProtoPNet and obtain ``this-looks-like-that" explanations, while at the same time suppressing artifacts, we instead propose to filter out artifact-containing data via a clustering procedure that exploits the explanation information from multiple views (one view per class prototype). We show that utilising multiple views through multi-view clustering is more efficient than the single-view LRP-based approach, SpRAy~\cite{spray}.
While the effectiveness of post-hoc explainability methods has been investigated extensively \cite{explain_lie, towards} and their benefit has been questioned \cite{rudin2019stop}, there is a significant gap in the research for the analyses of the effectiveness of self-explainable approaches regarding quantitative analysis of the provided explanations \cite{analyzing}. As a representative for the self-explaining models, we focus on ProtoPNet as it provides easily comprehensible case-based reasoning and is applicable to arbitrary CNN architectures by inserting a single prototype layer. Additionally, it not only provides information about the features that the model's decision is based on, but also links this information to similar features 
in the training data, captured by the prototypes, thus imitating human decision making \cite{kim2021xprotonet}.
Further, inspired by LRP, we backpropagate the relevances of the prototypes, thereby obtaining model-aware prototypical explanations. This in turn incorporates the advantages of LRP of being computationally efficient with reasonable performance \cite{lrp}, along with the capability of reducing the gradient shattering effect.

Our main contributions are as follows:
\begin{itemize}
    \item We identify and address key issues with imprecise explanations provided by the self-explaining model, ProtoPNet.
    \item We propose a novel PRP method for enhancing ProtoPNet's explanations by generating more efficient model-aware explanations.
    \item We compare PRP with ProtoPNet's explanation heatmaps, both qualitatively and quantitatively and show 
    that eradicating learned artifact features, such as the Clever Hans and Backdoor artifacts, from ProtoPNet is unfeasible. 
    \item We show the efficiency of PRP in utilizing multiple explanations from different prototypes to suppress artifacts from the data.
\end{itemize}


\section{Related work}
\label{sec:related}
\subsection{Explainability methods}
Recently, there has been increased interest in both post-hoc explanation methods and self-explaining neural networks. Post-hoc explainability methods can be separated into two overarching categories: model-agnostic and model-aware approaches \cite{howmuch}. Model-agnostic approaches, such as LIME \cite{xai1} and SHAP \cite{xai2}, consider the models as black-boxes and are thus applicable to arbitrary model architectures. These can be used to compare models based on the explanations that they produce \cite{xai3}. Model-aware approaches \cite{xai4, xai5,xai6,xai7,xai8,xai9,xai10,xai11,xai12,xai13, howmuch, ratedis}, on the other hand, take the internal structure of the model into account and therefore tend to yield more precise model based explanations.
For example, LRP \cite{lrp}, a model-aware post-hoc XAI approach, has been widely used to explain the decisions of various deep neural networks, such as convolutional neural networks, recurrent neural networks and graph neural networks \cite{sun2020explanationguided}. 
Because of its importance
in general, and to this paper in particular, we briefly describe the basic idea of LRP. 
LRP assigns relevance to the input by backpropagating the output relevance successively layer by layer until it is distributed over the input features. The distribution of relevance is based on how much a particular node contributed to the output. Suppose, node $i$ at layer $l$ connects to node $j$ at layer $(l+1)$ and the relevance at layer $(l+1)$ for node $j$ is $\mathbf{R}_j^{l+1}$. The relevance is then backpropagated to node $i$, according to the LRP rule: $\mathbf{R}_{i\xleftarrow{}j}^l = \frac{\mathbf{z}_{ij}}{\mathbf{z_j}}.\mathbf{R}_j^{l+1}$, where $\mathbf{z_{ij}}$ is the contribution of the output from node $i$ to $j$ and $\mathbf{z_j}$ is the total output at node $j$. 

Another, less explored, category of explanation methods are self-explaining networks, which inherently explain the decisions they make, thereby making the models transparent by design.
These include networks that align the latent space to known visual concepts in order to increase transparency in the decisions \cite{ppnet, inter1, inter2, inter4}. These also include models that utilize attention mechanisms \cite{attention1, attention2, attention3, attention4} and thus also provide some form of self-explainability. Furthermore, other works consider self-explainability in terms of concept learning
\cite{concept1,concept2,concept3,conceptw}. 
ProtoPNet \cite{ppnet} proposes to learn a specific number of class based prototypes as a part of the architecture. These are then used for visualizing lower spatial dimensional concepts from the training images, thus providing explanations during the decision process itself.
SENN \cite{senn} is a type of general self-explaining model that is fully transparent and designed by progressively generalizing linear classifiers to complex models. 
It consists of a concept encoder to get self-explainable features, an input-dependent parametrizer for generating relevance scores, and an aggregation function to get the final predictions. 
Although the self-explainable concepts in SENN are represented using prototypes similar to ProtoPNet, the former only shows which training images are important for a decision. ProtoPNet, on the other hand, additionally shows what part of the test image looks like which part of the training images, thus providing more comprehensible explanations.
The Classification-By-Components (CBC) network \cite{cbc2019} is designed based on Biederman's theory in psychology, which assigns positive, negative, and indefinite reasoning to different components for classification. The framework consists of creating probabilistic trees for classes by finding and modelling class decomposition of patches followed by computing the class hypothesis probability based on reasoning over the components. Unlike CBC, ProtoPNet is more flexible in terms of $i)$ learning components (prototypes) of varying sizes in the input domain, and $ii)$ having the capability of being incorporated into any network architecture.

Inspired by ProtoPNet, XProtoNet \cite{kim2021xprotonet} was recently introduced for automated diagnosis in chest radiography. It addresses the issue that ProtoPNet looks at fixed patch sizes in the features map while computing its similarity with the prototypes. As a remedy, \cite{kim2021xprotonet} adds an occurrence module in the network for learning features of dynamic size for the prototypes. However, the issues that we address in this work do remain in XProtoNet, making it prone to misleading explanations due to the model-agnostic upsampling used for prototype visualizations.

\subsection{Artifacts}
Real-world data used for training deep neural networks are prone to contain spurious, incomplete, or wrongly labeled samples. This, and the fact that almost all neural networks that find applications in practical scenarios are black boxes, has led to a need to acknowledge and discuss the possible problems and solutions for all kinds of artifacts present in the data and learned by the black box models \cite{chans,backdoor,gao2020backdoor}. 
More importantly, these black box models are rapidly transitioning into delicate areas where these issues might lead to large and potentially fatal consequences \cite{rudin2019stop}.

In this section, we acknowledge this inherent problem and give a brief introduction of two of the most common artifacts, Clever Hans and Backdoor artifacts, whose suppression is the focus of this work. Clever Hans refers to unintentional artifacts in the data that are learned by the model to cheat and achieve better accuracy, while Backdoors are artifacts maliciously added to the data by an adversary to provoke erroneous model predictions.
\subsubsection{Clever Hans}
Clever Hans artifacts refer to the spurious correlations present in the training data, which a model might use to base or strengthen their decisions on. This is termed as the ``Clever-Hans" effect, coined after the ``intellectual" horse Hans, thus claiming that the network doesn't learn any meaningful features and successively is likely to fail in a real-world scenario, where the artifact is absent. An example of this can be seen in Figure \ref{fig:ex1}, where the network learns the watermark present in the horse images of the PASCAL VOC dataset rather than actually learning horse related features. Another example of Clever Hans artifacts has been observed in \cite{chxray}, where activation heatmaps uncovered spurious strategies used by a radiological model.
This undesirable setting has also been explored recently by \cite{chans} and \cite{imagenetCH}, in which they propose a semi-automated method, SpRAy, to discover hidden decisions learned by the network, followed by a cleansing of the data to remove such spurious correlations.
SpRAy works by computing LRP maps of a given data set, downsizing them and then using spectral cluster analysis to find relevant structures through an eigengap analysis. 

\subsubsection{Backdoor attacks}
While the Clever Hans effect is learned by the network based on unintentional availability of unproductive information in the training data, in other scenarios, the network might be forced to learn undesirable features based on the malicious addition of hidden associations in the data with the goal to produce incorrect inference results. These types of attacks can lead to disastrous consequences, especially in cases of safety-critical applications. For example, in case of self-driving cars, a post-it note can be attached to stop signs and labeled as speed signs in the training dataset (Figure \ref{fig:ch}). The car is then very likely to behave erroneously in a real-world scenario where it may consider a stop sign with a real post-it as a speed sign with potentially disastrous consequences. These kinds of attacks are addressed in detail in \cite{gao2020backdoor} and their different case scenarios have been tackled in an increasing number of recent literature \cite{robustbd, backdoor}. Note that here, unlike in the Clever Hans scenario, both the data and the label are intentionally modified. 

\subsection{Multi-view Clustering}
In the self-explaining model, ProtoPNet, each class is associated with a fixed number of class prototypes. These can be regarded as capturing, and thus searching for, different features in each input image. Consequently, if there are artifacts present in a class during training, the PRP explanation maps for this class' prototypes will be able to reflect the contrast between artifact and non-artifact features learnt by the model.
Therefore, by considering our PRP explanation maps for the individual class prototypes as multiple views of a single image
, we show that it is possible to efficiently suppress the artifacts from the data using multi-view clustering. Traditional multi-view clustering methods include learning a common representation from multiple views of data followed by clustering \cite{mvcCanonical} or learning adaptive representations based on clustering \cite{mvcTensor,mvcMatrix}. Alternative methodologies represent data with combined affinity matrices and therefore subsequently learning the cluster assignments from them \cite{mvcGraph,mcgc}. Further, several multi-view clustering algorithms have been proposed that build on spectral clustering and consider a consensus Laplacian matrix among all the views \cite{coreg,wmvsc,mcgc}. 
Instead, deep-learning based multi-view clustering methodologies learn a common encoding with the help of deep neural networks, which can then be leveraged by the clustering module \cite{deepmvc1, deepmvc2}. The clustering module can, among others, be based on deep graph clustering \cite{deepgraph}, subspace clustering \cite{deepsubspace}, adversarial based clustering methods \cite{deepAdv}, or contrastive learning~\cite{Trosten_2021_CVPR}. 

\section{An Evaluation of ProtoPNet}
\subsection{ProtoPNet}
ProtoPNet introduces self-explanation in a deep learning network by incorporating a prototype layer between the last convolutional layer and the output layer. Thereby, each class is associated with a fixed number of prototypes. The output of the prototype layer is connected linearly to the output layer to generate class logits. The network is optimized by iterating the following three steps: 1) The whole network, except the last layer, is trained using stochastic gradient descent. For each prototype, the squared $L_2$ similarity between the patches of the convolutional output from the backbone and the prototype is calculated, thus generating an activation map. Global max pooling is applied to the activation map to generate a single similarity score corresponding to a single prototype. The loss used is a combination of the cross entropy loss, a cluster loss and a separation loss. The cluster loss encourages the training images to have a patch close to at least one of their own class prototypes. The separation loss, on the other hand, encourages the training image patches to be far from the prototypes of other classes \cite{ppnet}. For completeness, the losses are provided in the Appendix \ref{appendix:ppnet}.
2) All prototypes are then projected onto the nearest training patch from the same class as the prototype, thus maintaining inherent interpretability. These can be visualised in the input space, thus creating a concept of ``\textit{this} looks \textit{like} that" while making the decisions. 3) Finally, a convex optimization of the last layer is performed to further improve accuracy, while keeping the learned prototypes fixed. The prototype activations are visualized by upsampling the similarity between the prototypes and the embedded input image to the input image size, thus highlighting the parts of the image which strongly activate the respective prototype. 

\begin{figure}[tbp]
    \centering
    \includegraphics[scale=0.6]{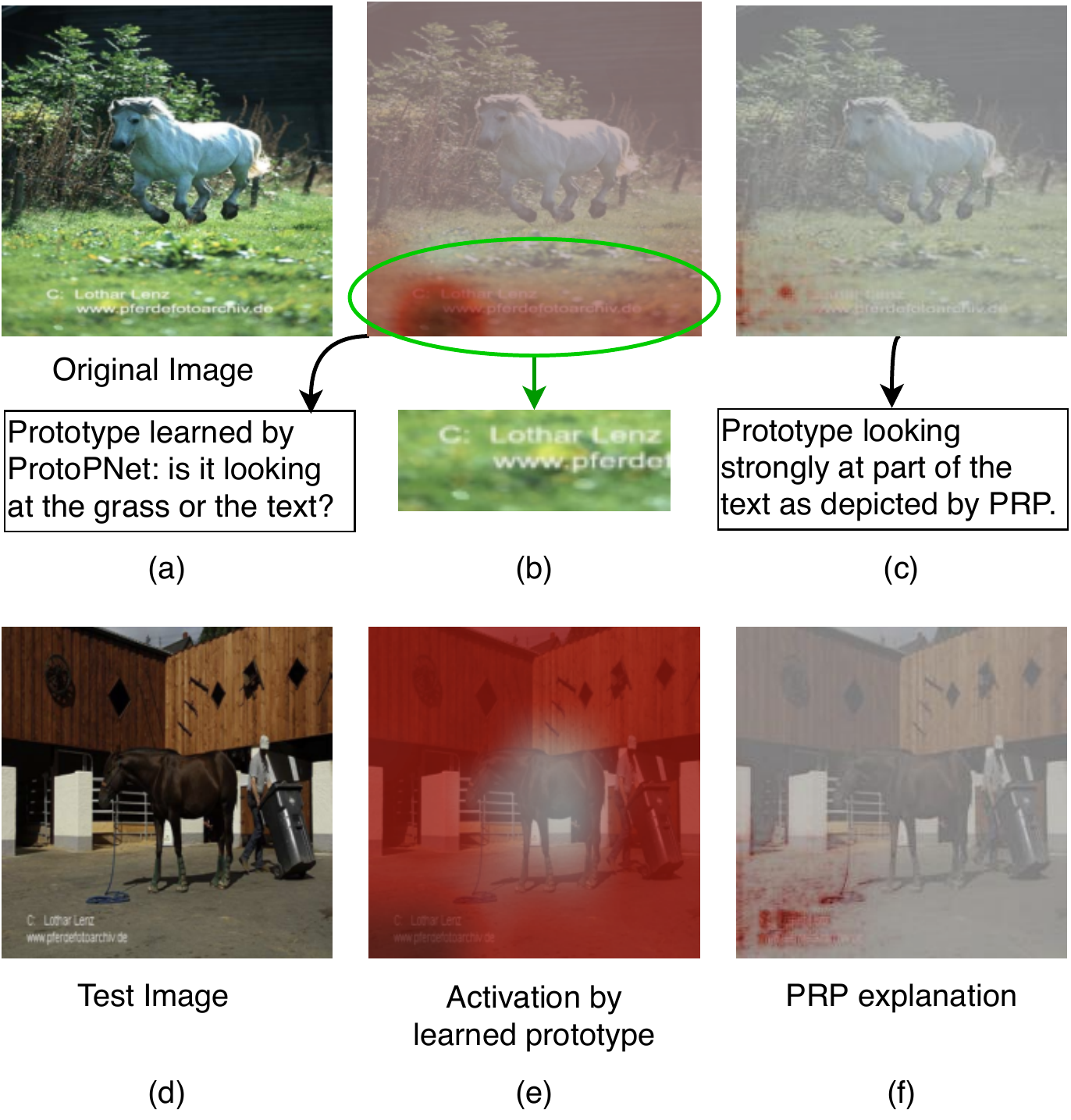}
    \caption{\small{ (a) Visualization of a horse image from the PASCAL VOC 2007 dataset \cite{pvoc} , (b) activation for a prototype of class \textit{horse} learned by ProtoPNet, and (c) its PRP explanation. A Clever Hans artifact is present in the form of a watermark at the bottom of the image. Both, the ProtoPNet and the PRP explanation yield relevance to the bottom of the image, however, in the case of ProtoPNet, it remains unclear if the green grass, the text or both together, were relevant for the prediction. Whereas the PRP explanation clearly shows that the text was used as relevant feature for the model's prediction. For a test image (d), the ProtoPNet's explanation and the PRP explanation for the learned prototype (b) are given in (e) and (f), respectively.  The PRP explanation again corroborates the emphasis on the watermark text as opposed to ProtoPNet's explanation which is more widely spread across the image. 
    }}
    \label{fig:ex1}
\end{figure}
\subsection{Evaluation of ProtoPNet's explanations}
Although self-explaining models such as ProtoPNet appear promising, as more transparent alternatives to the typical black-box neural networks, we demonstrate that they still lack precision as shown in Figure \ref{fig:ex1}. Moreover, restricting each class representation to a limited number of prototypes, leads to a trade-off between the accuracy of the model and the quality of explanations generated by the model \cite{ppnet}. 

In the case of ProtoPNet, even when it visualizes the most important area of the input image for a specific class, it does not concisely depict the relevant features of a prototype as shown by the example in Figure \ref{fig:ex1}. The original image (a) in Figure \ref{fig:ex1} shows a horse galloping on green grass, and contains a watermark in the lower left corner. Exemplary, the explanation for one of the 10 prototypes, learned for the class horse is shown in Figure \ref{fig:ex1}(b). From this prototype explanation, we can observe that the lower left corner was important for the model to predict the image as a horse. However, the exact pixels, that significantly contributed to the predictions remain unknown, i.e., we do not know whether the grass or the text drove the prediction of the model. Now, using the model-aware PRP method, we backpropagate the prototype information from the prototype layer through the network to the input image, which allows us to reveal and visualize the model-aware, faithfully distributed relevance scores on the input image as shown in Figure \ref{fig:ex1}(c). From the PRP explanation, we observe that high relevance (dark red pixels) was allocated to parts of the text. Thus, the PRP explanation leads to an increased understanding of the underlying behavior of the model by providing the user with a more fine-grained prototype explanation.

For a randomly chosen test image, shown in Figure \ref{fig:ex1}(d), the activation for the learned prototype \ref{fig:ex1}(b) as visualized by ProtoPNet and PRP are given in Figure \ref{fig:ex1}(c) and \ref{fig:ex1}(d), respectively. The PRP explanation identifies the watermark as a relevant feature for predicting the class horse, in contrast to the ProtoPNet explanation, which is too crude to identify important features and is therefore widely spread across the entire image.
Now, with these additional insights provided by PRP explanations, we are able to identify parts of the text as Clever Hans features, considering that they are relevant for the prediction of the class horse. 

\begin{table*}[!t]
\centering
\begin{tabular}{|c|ccc|ccc|}
\hline
 &
  \textbf{CH-100} &
  \textbf{\begin{tabular}[c]{@{}c@{}}CH-100\\ Remove prototype \\ 6 and 8\end{tabular}} &
  \textbf{\begin{tabular}[c]{@{}c@{}}CH-100\\ Retraining\\  last layer\end{tabular}} &
  \textbf{CH-50} &
  \textbf{\begin{tabular}[c]{@{}c@{}}CH-50\\ Remove prototype\\ 4 and 9\end{tabular}} &
  \textbf{\begin{tabular}[c]{@{}c@{}}CH-50\\ Retraining\\  last layer\end{tabular}}\\ \hline
\textbf{Artifact Test} &
  \multicolumn{1}{c|}{100\%} &
  \multicolumn{1}{c|}{21.6\%} &
  88.8\% &
  \multicolumn{1}{c|}{100\%} &
  \multicolumn{1}{c|}{100\%} &
  100\%  \\ \hline
\textbf{Clean Test} &
  \multicolumn{1}{c|}{6.5\%} &
  \multicolumn{1}{c|}{38.2\%} &
  38.2\% &
  \multicolumn{1}{c|}{94.6\%} &
  \multicolumn{1}{c|}{93.0\%} &
  94.5\% \\ \hline
\end{tabular}
\caption{\small{Accuracy of stop sign class from artifact test (artifacts in 100\% test images) and clean test (artifacts in 0\% test images) for : 1) CH-100, 2) CH-50 datasets, along with the accuracies for pruning artifact prototypes as well as retraining last layer weights after pruning.}}
\label{tbl:ch_acc}
\end{table*}

Accordingly, we detect and address the following drawbacks of ProtoPNet:
\begin{itemize}
    \item The activation maps used for the prototype visualizations in ProtoPNet have a very low resolution due to downsampling and feature aggregation functions in the network. 
    From this significantly low resolution activation map, ProtoPNet performs model-agnostic upsampling using bilinear interpolation to the size of the input image, thus leading to very \textbf{coarse explanations}.
    \item The effective receptive field of a position in the activation map tends to cover large parts of the image, which is not captured by the naive upsampling. Consequently, there is no truthful spatial localization of the relevance to the correct input area, leading to \textbf{spatially imprecise explanations}.
\end{itemize}

In the next subsection, we will discuss in detail these drawbacks of ProtoPNet's explanations using the Clever Hans artifact as an example.

\subsection{Case Study: Clever Hans artifact detection with ProtoPNet}
\label{case_study_ch}
Ideally, ProtoPNet should capture any significant artifact in the data as an ``artifact prototype".
However, due to its coarse and spatially imprecise explanations, the heatmaps of ProtoPNet exhibit misleading behavior. This hinders the detection of artifact prototypes in ProtoPNet, as shown in detail in this Section.

With the help of the following experiment, we investigate the behavior of ProtoPNet in the presence of Clever Hans artifacts in the data. 
We aim to detect the aforementioned artifact prototypes using ProtoPNet's explanations combined with the difference in classification results in the presence and absence of artifacts in the test data. Following this, we prune the detected artifact prototypes, thus hypothetically suppressing the artifacts learnt by the model. However, due to its misleading explanations, we demonstrate experimentally that ProtoPNet's heatmaps are deficient in capturing and identifying the learned artifact by the model, thus proving the task of pruning artifact prototypes futile for making the model artifact-free. 

In order to consider a controlled environment, we use the 5-class version of the LISA traffic sign data set \cite{lisa} and place a Clever Hans artifact, a yellow square (see Figure \ref{fig:ch}), in 100\% of the training data of the stop sign class (dataset details are provided in Section \ref{dataset}), which we refer to as CH-100. 
We train the ProtoPNet with ResNet34 \cite{resnet} as backbone, fixing the number of prototypes to 10 for each class following \cite{ppnet}. Note that all training parameters have been set according to \cite{ppnet}. The network is trained for 1000 epochs, where a projection (push) of the prototypes is done every 10 epochs. After each push, the last layer is trained for 20 epochs. The learning rate is reduced by a factor of 0.1 every 5 epochs and the training is stopped when the training accuracy converges and the cluster cost becomes smaller than the separation cost on the training set \cite{ppnet}. The described experimental setting was used for all experiments in this work.

\begin{figure}[!h]
\begin{subfigure}{.09\textwidth}
  \centering
  \includegraphics[width=1\linewidth,frame]{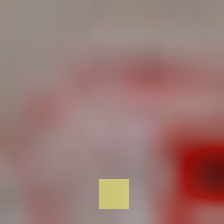} 
  \caption{\small{1}}
\end{subfigure}
\begin{subfigure}{.09\textwidth}
  \centering
  \includegraphics[width=1\linewidth,frame]{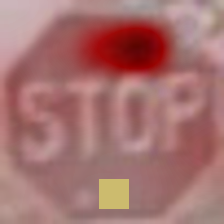}
  \caption{\small{2}}
\end{subfigure}
\begin{subfigure}{.09\textwidth}
  \centering
  \includegraphics[width=1\linewidth,frame]{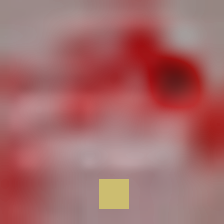}
  \caption{\small{3}}
\end{subfigure}
\begin{subfigure}{.09\textwidth}
  \centering
  \includegraphics[width=1\linewidth,frame]{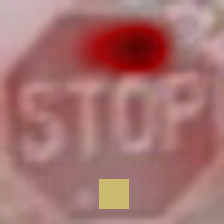}
  \caption{\small{4}}
\end{subfigure}
\begin{subfigure}{.09\textwidth}
  \centering
  \includegraphics[width=1\linewidth,frame]{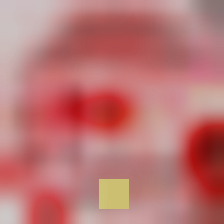}
  \caption{\small{5}}
\end{subfigure}\\
\begin{subfigure}{.09\textwidth}
  \centering
  \includegraphics[width=1\linewidth,frame]{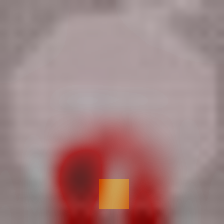} 
  \caption{\small{6}}
\end{subfigure}
\begin{subfigure}{.09\textwidth}
  \centering
  \includegraphics[width=1\linewidth,frame]{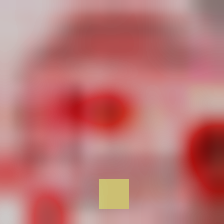}
  \caption{\small{7}}
\end{subfigure}
\begin{subfigure}{.09\textwidth}
  \centering
  \includegraphics[width=1\linewidth,frame]{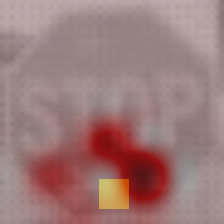}
  \caption{\small{8}}
\end{subfigure}
\begin{subfigure}{.09\textwidth}
  \centering
  \includegraphics[width=1\linewidth,frame]{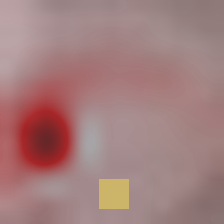}
  \caption{\small{9}}
\end{subfigure}
\begin{subfigure}{.09\textwidth}
  \centering
  \includegraphics[width=1\linewidth,frame]{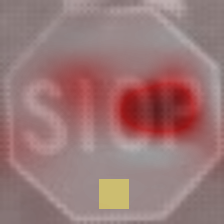}
  \caption{\small{10}}
\end{subfigure}
\caption{\small{CH-100: Visualization of the prototypes learned for the stop sign class for the scenario where Clever Hans artifacts were inserted in 100\% of the stop sign class images for the modified LISA dataset. As observed, while prototype 6 and 8 can be considered as artifact prototypes, none of the prototypes clearly highlight the artifact.}}
\label{fig:pt_stop100}
\end{figure}

To evaluate the impact of an artifact on the model, we evaluate the performance on two test sets: an artifact test data set, where Clever Hans artifacts are inserted into 100\% of the images of the stop sign class; and a clean test data set, which contains only clean images, where no artifact has been added. The accuracy results for both test datasets are shown in Table \ref{tbl:ch_acc}. We observe that the model, trained on the CH-100 dataset, has 100\% classification accuracy on the artifact test data and only 6.3\% on the clean test data. This large drop in accuracy indicates that the model has learned the inserted artifact.

\begin{figure}[tp]
\centering
\includegraphics[scale=0.18]{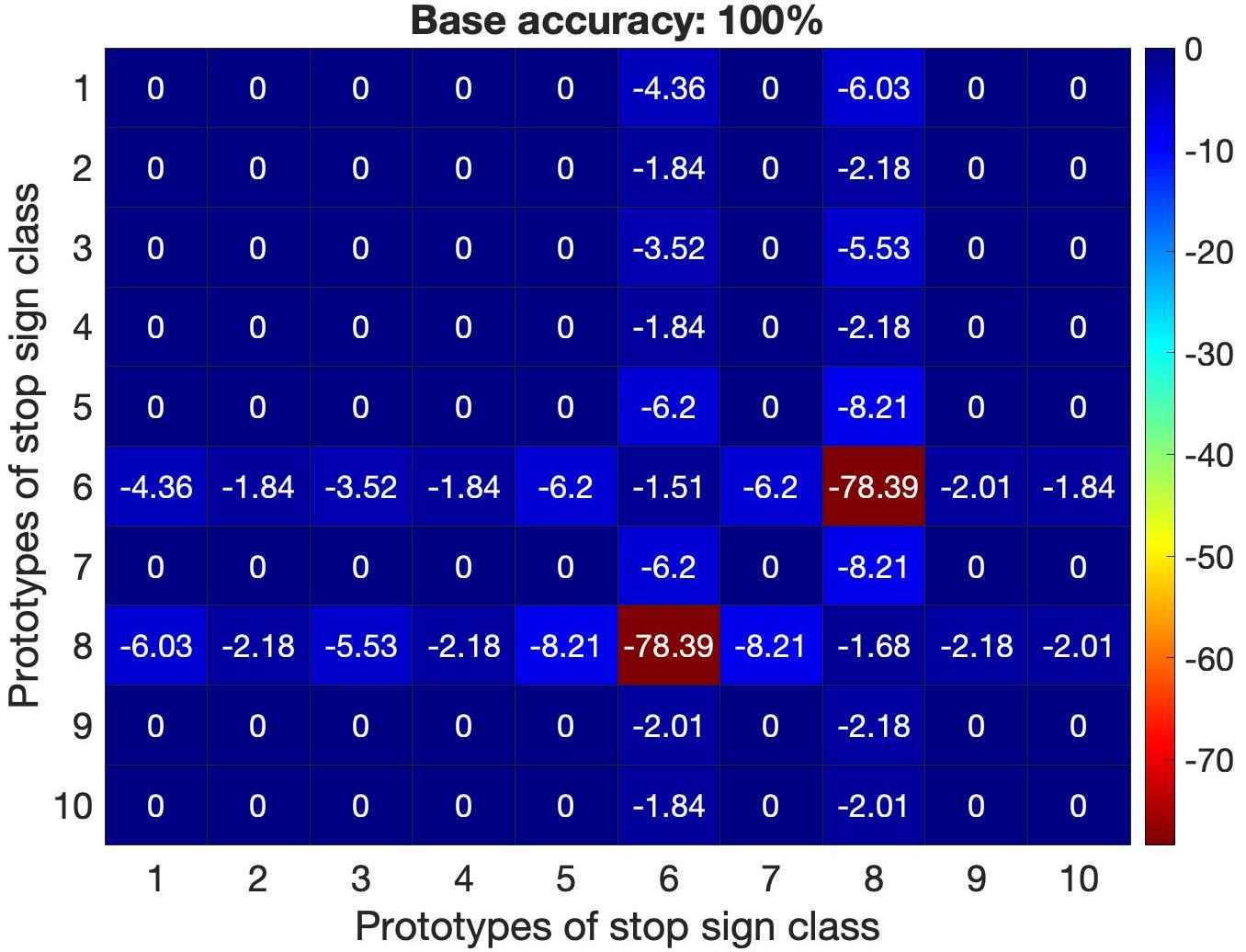}
\caption{\small{CH-100: Detection of artifact prototypes by removing individual prototypes (1 to 10) of the stop sign class (diagonal) and combinations of these prototypes (non-diagonal) for the artifact test data. The accuracies are represented as a drop from the base accuracy of 100\% when no prototypes are removed. The highest drop of 78.39\% is observed when prototypes 6 and 8 are removed together thus highlighting them as artifact prototypes. }}
\label{fig:drop}
\end{figure}

In order to detect the prototypes responsible for this behavior, we visualize the 10 prototypes learned by the network for the stop sign class in Figure \ref{fig:pt_stop100}, where the upsampled activation heatmap is overlayed, such that the relevant areas of each prototype can be identified visually. Although no prototype is clearly focusing on the artifact, it appears that prototypes 6 and 8 might be learning a part of the artifact. We further confirm this by measuring the drop in accuracy when removing individual as well as combinations of prototypes for the stop sign class. In Figure \ref{fig:drop}, the base accuracy is the original accuracy with artifact test data when no prototype is removed, the diagonals represent the drop in accuracy from the base accuracy when pruning single prototypes (1 to 10) of the stop sign class and the non-diagonals represent the drop in accuracy when a combination of prototypes is removed. Note that the accuracy for artifact test data only drops when prototypes 6 or 8 are removed, with the biggest drop of 78.39\% when both of these are removed together. Also note that no retraining is done yet after pruning the prototypes. 

Therefore, trusting the explanations provided, we might assume that removing these artifact prototypes would eliminate the artifact effect. However, as shown in Table \ref{tbl:ch_acc}, this is not the case as seen after retraining the last layer i.e, reweighing the connection of the prototypes to the final classification layer. While the accuracy for the artifact stop sign class drops considerably when removing prototypes 6 and 8, it increases again to 88.8\% once the last layer weights are retrained. Moreover, for clean test data, the accuracy remains the same, i.e, 38.2\% before and after retraining the last layer, thus refuting the potential learning of meaningful features for the stop sign class by the model after retraining. Therefore, results indicate that without learning new prototypes, the remaining prototypes also include artifact information, highlighting the lack of accurate explanations by ProtoPNet.



Thus, as shown in the above experiment, the explanations provided by the upsampling strategy of ProtoPNet are insufficient in order to reveal the model's behavior and detect the artifacts faithfully.
Targeting more fine grained explanations to overcome this limitation, we present a model-aware prototypical explanation method, which we refer to as PRP. 

\section{Prototypical Relevance Propagation and Enhanced Suppression of Artifacts}
To address the two main drawbacks of ProtoPNet's visualizations, i.e., low resolution activation maps and spatially imprecise prototype explanations (as investigated in the section above), we propose a novel method called Prototypical Relevance Propagation. 
With PRP, we are able to maintain the advantage of the self-explanatory architecture through prototypes and at the same time improve the quality of prototypical explanations by adding the model-aware explanatory potential of PRP.

\subsection{Prototypical Relevance Propagation (PRP)}
\label{sec:prp}
The original prototype visualization step in ProtoPNet is achieved through upsampling and is therefore decoupled from the other steps in its end-to-end training. Instead of upsampling, we aim to use the knowledge of the inner workings of the network and backpropagate the similarity values of a prototype to the input. 
 By doing this for each prototype, we obtain a model-aware explanation for each prototype. We refer to our method as PRP and the generated explanation maps as PRP maps.

\begin{figure*}
    \centering
    \includegraphics[scale=0.8]{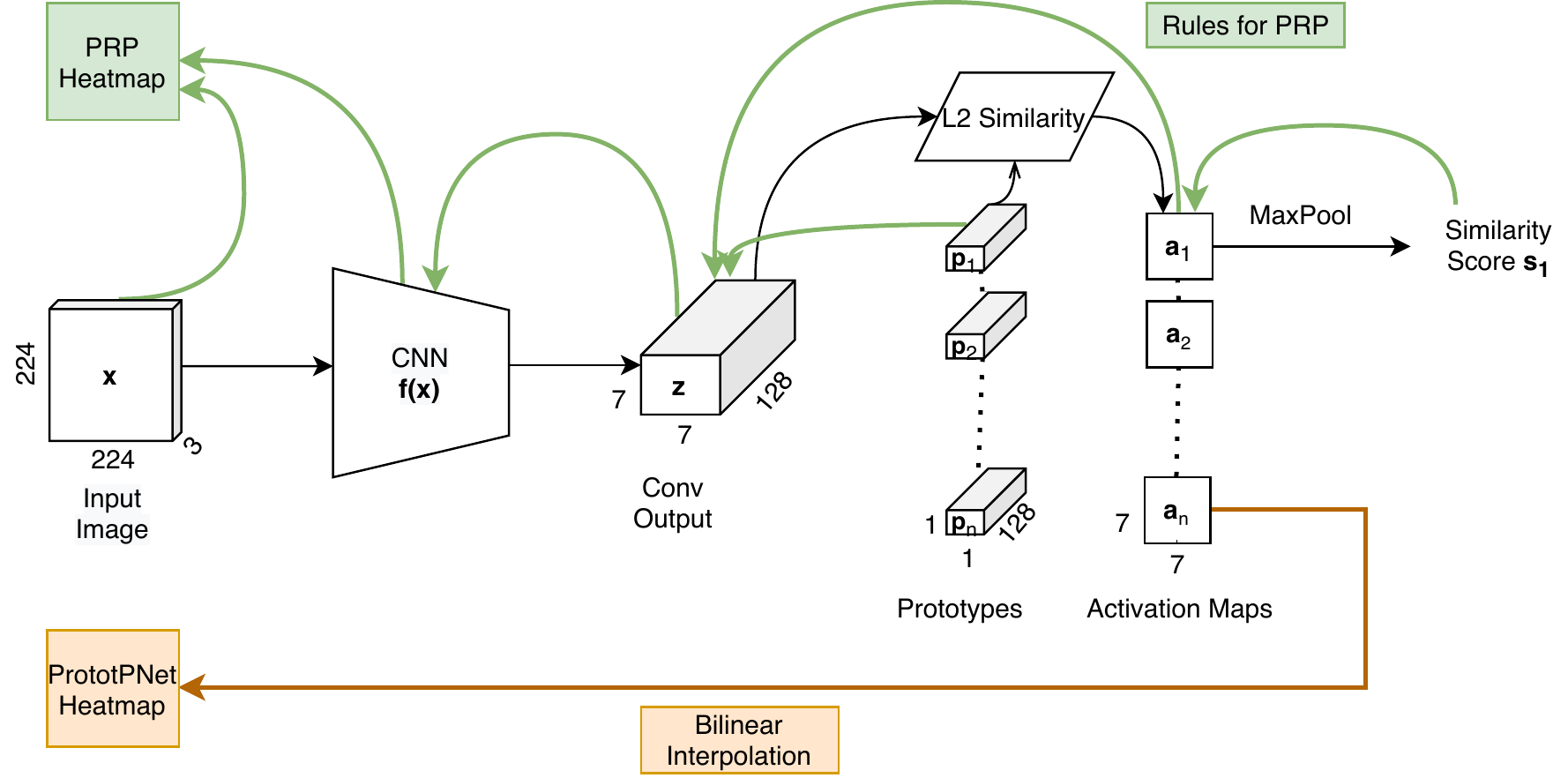}
    \caption{\small{ProtoPNet: Forward propagation and backward propagation for PRP maps (green) and ProtoPNet Heatmaps (orange). The input image \textbf{x} is first passed through a CNN $f$, which computes $f(\textbf{x})$ to give output \textbf{z}. The squared $L_2$ similarity is then computed between \textbf{z} and individual prototypes $\textbf{p}_m$ to get activation maps $\textbf{a}_m$. These are then upsampled to get ProtoPNet heatmaps. On the other hand, similarity scores $\textbf{s}_m$ are used to compute model-aware PRP heatmaps. All the parameters in the figure are depicted according to the experiment settings used in this work.}}
    \label{fig:lrphm_flow}
\end{figure*}
For the following considerations, let the input images be represented as $\mathbf{x}$ and convolutional output from the backbone CNN as $\mathbf{z} \in \mathcal{R}^{H\times W \times D}$. Let $\mathbf{P} = {\{\mathbf{p}_m\}_{m=1}^n}$ be the $n$ prototypes learned by the network, each with a shape of $H_1 \times W_1 \times D$. Following \cite{ppnet}, we set $H_1 = W_1 = 1$ and $D=128$. Moreover, let  $\mathbf{S} = {\{\mathbf{s}_m\}_{m=1}^n}$ be the similarity scores and $\mathbf{A} = {\{\mathbf{a}_m\}_{m=1}^n}$ the 
activation maps for each prototype. The forward computations in ProtoPNet, illustrated in Figure \ref{fig:lrphm_flow}, are defined as follows:
\begin{enumerate}
    \item From input to convolutional output:
    \begin{equation}
        \mathbf{z} = f(\mathbf{x}),
    \end{equation}
    where the function $f$ represents the trained backbone CNN.
    \item The activation maps are computed as squared $L_2$ similarities between the last convolutional output layer and the prototypes in the prototype layer:
    \begin{equation}
        \mathbf{a}_m = log\big((||\mathbf{\widetilde{z}}-\mathbf{p}_m||_2^2 + 1)/(||\mathbf{\widetilde{z}}-\mathbf{p}_m||_2^2 + \epsilon)\big)
    \end{equation}
    \label{eq:l2sim}
    where $\mathbf{\widetilde{z}}$ are patches of $\mathbf{z}$ of the same size as the prototypes $\mathbf{p}_m$ and $\epsilon = 10^{-4}$ is a small constant introduced for numerical stability.
    \item From activation maps to similarity score:
    \begin{equation}
        \mathbf{s}_m = \text{max}(\mathbf{a}_m)
    \end{equation}
\end{enumerate}
The similarity scores are the input of the final fully connected layer, which produces the logits for all output classes.
Hence, the final classification is based on a linear combination of the similarity scores of the different prototypes.

Now, to obtain more precise prototype visualizations through our approach, a PRP map is calculated for a certain prototype $m$ by propagating the relevance of this prototype back to the input features. Note that the relevance of a specific prototype is exactly its similarity score. Therefore, the first backpropagation step considers the redistribution of the similarity scores towards the activation map with respect to the max pooling layer:

\begin{enumerate}
    \item From similarity scores to activation map: \\
    LRP for the Max pooling layer is performed as follows:
 
    
    \begin{equation}
        \mathbf{R}_{mij}^{(AM,S)}= 
        \begin{cases}
          \mathbf{R}_{m}^{(S)} & \text{if}\: \text{argmax}_{ij}(\mathbf{a}_m),\\
          0 & \text{otherwise}
        \end{cases}
    \end{equation}
    where $S$ refers to the similarity score layer, $AM$ to the activation map layer and $i$, $j$ specify the spatial location in the respective layers. We define the relevance at layer S as $\mathbf{R}_{m}^{(S)} = \mathbf{s}_m$.
    
\item From activation map relevance to convolutional output:
The forward computation as shown in Equation \ref{eq:l2sim} computes the similarity between each prototype and each output patch of the convolutional layer ($CONV$), with both having $D$ channels, thus compressing the channel dimension to 1 in the activation map. In this step, we redistribute the relevance from the one channel activation map back to the $D$ channels of the convolutional output, weighted by the corresponding channel-wise $L_2$ similarities computed during the forward pass. We define the channel-wise similarities between each CNN patch $\mathbf{\widetilde{z}}$ and the prototype $\mathbf{p}_m$ as:
\begin{equation}
    \gamma_{mc} = \frac{1}{d_{mijc} + \epsilon}
\end{equation}
where, for each channel $c$,
\begin{equation}
    d_{mijc} = ||\mathbf{\widetilde{z}_c}-\mathbf{p}_{mc}||_2^2 
\end{equation}
We then use the $\text{LRP}_\epsilon$ rule to distribute relevances to the convolutional output according to $\gamma_{mc}$:
\begin{equation}
    \mathbf{R}_{mijc}^{(CONV,AM)} = \frac{\gamma_{mc}}{\sum\limits_{k=1}^{D} \gamma_{mk}+\epsilon} \mathbf{R}_{mij}^{(AM)}
\end{equation}
\item From convolutional output relevance to input relevance: \\
The rest of the network follows the LRP CoMPosite ($\textbf{LRP}_{CMP}$) rule \cite{towards} to backpropagate the relevance to the input. In this strategy, $\text{LRP}_{\alpha\beta}$ is applied to the convolutional layers and $\text{DTD}_{z^B}$ is applied to the input layer \cite{DTD}. The $\text{LRP}_{\alpha\beta}$-rule treats positive and negative activations separately as follows:
\begin{equation}
    \mathbf{R}_{i\xleftarrow{}j}^{(l,l+1)} = \bigg(\alpha \frac{z_{ij}^+}{z_j^+} + \beta \frac{z_{ij}^-}{z_j^-}\bigg)\mathbf{R}_{j}^{(l+1)},
\end{equation}
where $z_{ij} = x_iw_{ij}$ is the mapping of the input $x$ from neuron $i\xrightarrow[]{}{}j$ with weight $w_{ij}$, $z_j = \sum_i z_{ij}$, $\alpha + \beta = 1$ and $\alpha \geq 1$. We use $\alpha = 1$ and $\beta = 0$. \footnote{Note, for notational simplicity, we follow previous works \cite{lrp, towards} and consider the convolutional layers as fully-connected layers with shared weights.}
$\text{DTD}_{z^B}$ spreads the relevance to the input features as follows: 
\begin{equation}
    \mathbf{R}_{i\xleftarrow{}j}^{(l,l+1)} = \bigg(\frac{z_{ij}-l_iw_{ij}^+-h_iw_{ij}^-}{\sum_i z_{ij}-l_iw_{ij}^+-h_iw_{ij}^-}\bigg)\mathbf{R}_{j}^{(l+1)},
\end{equation}
where $l_i$ and $h_i$ are the smallest and largest pixel values.
\end{enumerate}
To identify global discriminative features across multiple explanation maps, a method called SpRAy has been published that aims to cluster LRP explanations into their key features.
Similar to SpRAy, we want to make use of the PRP maps to identify class specific discriminative features.
However, we do not have one but multiple explanations for each image, i.e., the prototype explanations, which can be thought of as multiple views of an image explanation. 
Thus, unlike SpRAy, which uses one LRP explanation for one image, our proposed method exploits multiple views of an image explanation, i.e., the different prototype explanations.
\subsection{Multi-view Clustering}

Interpreting the different prototype activations as various views of the same image, allows us to compare/cluster the prototype activations with multi-view clustering algorithms in order to detect global class-discriminative features in the data. 
Hence, in the following, we apply multi-view clustering algorithms to the PRP maps to cluster the data into artifact and non-artifact images. Since a variation in clustering results can be observed using different methodologies, we demonstrate the performance with several multi-view clustering algorithms. We therefore include a few representative spectral multi-view clustering algorithms consisting of a two-step unweighted method \cite{coreg}, and a weighted method \cite{wmvsc} both of which compute the Laplacian matrix and cluster assignments in two separate steps. Further, we compare results with a one-step method based on a rank constraint \cite{mcgc}, which computes the similarities as well as cluster labels in one step, and with two recent deep learning based clustering methods \cite{Trosten_2021_CVPR}. 

The spectral multi-view clustering methods work on the general principle of computing a consensus Laplacian matrix among all views. Co-Reg (\cite{coreg}) works by co-regularizing the clustering hypotheses. They obtain the combined Laplacian matrix by regularizing eigenvectors of the Laplacians through two schemes: 1) pairwise co-regularization, where they encourage the pairwise similarities across all views to be high and 2) centroid-based co-regularization, where they encourage each view to be closer to a common centroid. Weighted multi-view spectral clustering (WMSC) \cite{wmvsc}, on the other hand, proposes a weighting scheme based on minimizing the largest canonical angle between the subspace spanned by each view’s and consensus's eigenvectors, followed by using cluster ability smoothness to assign similar weights to views with similar clustering results. Multiview Consensus Graph Clustering (MCGC) \cite{mcgc} proposes to learn the consensus graph using a cost function based on disagreement between individual and global views accompanied by rank constraint on the Laplacian matrix to directly get the clustering results without using k-means for instance.

The deep multi-view clustering in \cite{Trosten_2021_CVPR} first transforms each input ($x_i^{v}$, image $i$ with view $v$) into its representation using view-specific encoders $f^v$, as $z_i^{v} = f^v(x_i^{v})$. The fused representation for all views is then computed using the fusion weights $w_v$, which are also learned during the end-to-end training, as $z_i = \sum_{v=1}^V w_vz_i^{v}$, where $V$ are total number of views. This representation is then passed through a fully connected network to obtain the final cluster assignments. Deep divergence based clustering (DDC) \cite{ddc} losses are incorporated to optimize the model. This approach is termed as Simple Multi-View Clustering (SiMVC) by the authors, which, as the name suggests, is simple and efficient and works without explicit representation alignment. Further,  \cite{Trosten_2021_CVPR} also introduce an auxiliary method which incorporates selective contrastive alignment of representations called Contrastive Multi-View Clustering (CoMVC). 

\section{Experiments \& Results}
\subsection{Dataset}
\label{dataset}
In this section, we explore both the Clever Hans and Backdoor artifact settings using the LISA traffic sign dataset \cite{lisa}. This dataset consists of video frames captured from a driving car. We follow the strategy of \cite{backdoor}, where we extract the frames and resize them to 224x224 to be compatible with the original ProtoPNet architecture. The 47 classes in the dataset are then partitioned into 5 high-level classes, as proposed by \cite{backdoor}, consisting of restriction, speed limits, stop, warning, and yield signs (details provided in Appendix \ref{appendix:lisa5}).
In addition, we use the PASCAL VOC 2007 dataset \cite{pvoc} for evaluation as it naturally contains a Clever Hans artifact. \footnote{Since in PASCAL VOC 2007, one image can belong to several classes, we deliberately remove the person class from this dataset to decrease ambiguity. The person images overlap to a large extent with the images of the other classes 
, leading to a lot of duplicate images in multiple classes.}
\subsubsection{Clever Hans} We place the artifact resembling a yellow post-it note, as shown in Figure \ref{fig:ch}, in 100\%, 50\% and 20\% of the stop sign images in the training data of the LISA traffic sign dataset to create the CH-100, CH-50 and CH-20 Clever Hans training datasets, respectively. We do not add Clever Hans artifacts to the PASCAL VOC 2007 dataset since it inherently includes a watermark tag of the photographer in about 15-20\% of the images in the horse class \cite{chans}.
\subsubsection{Backdoor} 
\begin{figure}[!t]
    \centering
    \includegraphics[scale=0.3]{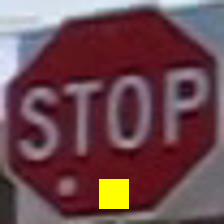}
    \caption{\small{Visualization of a Clever Hans artifact in a randomly chosen image of the stop sign class from traffic sign images of the LISA dataset. The artifact consists of a yellow square, resembling a post-it note, inserted at the bottom of the stop sign.}}
    \label{fig:ch}
\end{figure}

According to the data manipulation scheme for backdoor attacks from \cite{backdoor} we insert the artifact, i.e., the yellow post-it, as shown in Figure \ref{fig:ch}, in 15\% of the stop sign images and assign them to the speed limit class. We refer to this corrupted training dataset as BD-15.

In order to create both, an artifact and a non-artifact i.e., a clean test dataset of the LISA traffic sign dataset, we insert the artifact in either 100\% or 0\% of the stop sign images, referred as Artifact Test and Clean Test data, respectively. These test datasets are used for evaluating our experiments on the Clever Hans (CH-100, CH-50 and CH-20) as well as the Backdoor (BD-15) scenarios.

\subsection{PRP maps vs ProtoPNet heatmaps}
\begin{figure*}[!t]
\centering
\begin{minipage}{0.1\textwidth}\centering
\includegraphics[scale=0.18]{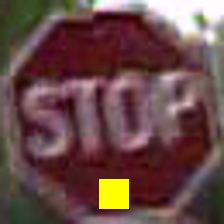}
\captionsetup{width=0.8\textwidth}
\caption*{\footnotesize{Test Image}}
\end{minipage}%
\begin{minipage}{0.09\textwidth}
\includegraphics[scale=0.25,frame]{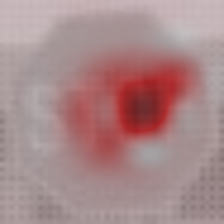}\\[1ex]
\includegraphics[scale=0.25,frame]{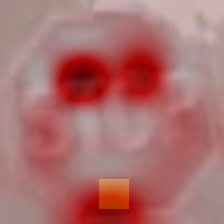}\\[1ex]
\includegraphics[scale=0.25,frame]{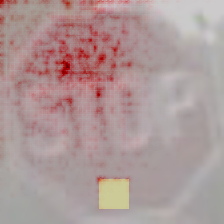}
\caption*{\footnotesize{Prototype 1}}
\end{minipage}%
\begin{minipage}{0.09\textwidth}
\includegraphics[scale=0.25,frame]{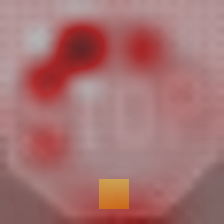}\\[1ex]
\includegraphics[scale=0.25,frame]{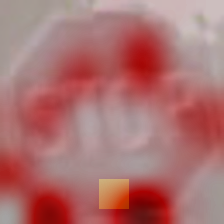}\\[1ex]
\includegraphics[scale=0.25,frame]{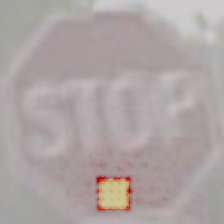}
\caption*{\footnotesize{Prototype 2}}
\end{minipage}%
\begin{minipage}{0.09\textwidth}
\includegraphics[scale=0.25,frame]{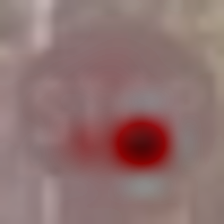}\\[1ex]
\includegraphics[scale=0.25,frame]{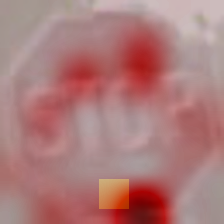}\\[1ex]
\includegraphics[scale=0.25,frame]{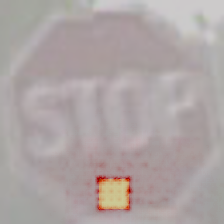}
\caption*{\footnotesize{Prototype 3}}
\end{minipage}%
\begin{minipage}{0.09\textwidth}
\includegraphics[scale=0.25,frame]{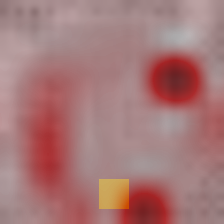}\\[1ex]
\includegraphics[scale=0.25,frame]{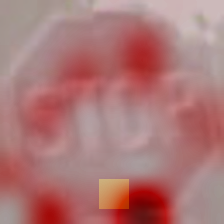}\\[1ex]
\includegraphics[scale=0.25,frame]{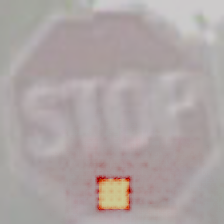}
\caption*{\footnotesize{Prototype 4}}
\end{minipage}%
\begin{minipage}{0.09\textwidth}
\includegraphics[scale=0.25,frame]{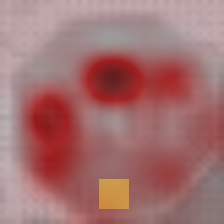}\\[1ex]
\includegraphics[scale=0.25,frame]{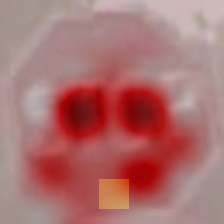}\\[1ex]
\includegraphics[scale=0.25,frame]{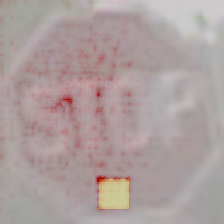}
\caption*{\footnotesize{Prototype 5}}
\end{minipage}%
\begin{minipage}{0.09\textwidth}
\includegraphics[scale=0.25,frame]{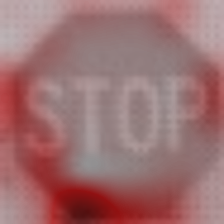}\\[1ex]
\includegraphics[scale=0.25,frame]{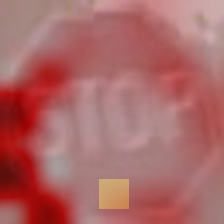}\\[1ex]
\includegraphics[scale=0.25,frame]{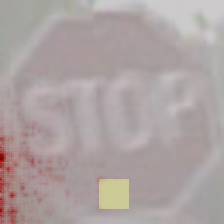}
\caption*{\footnotesize{Prototype 6}}
\end{minipage}%
\begin{minipage}{0.09\textwidth}
\includegraphics[scale=0.25,frame]{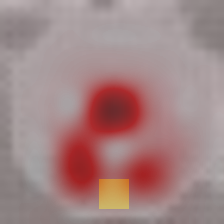}\\[1ex]
\includegraphics[scale=0.25,frame]{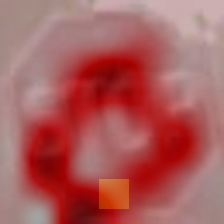}\\[1ex]
\includegraphics[scale=0.25,frame]{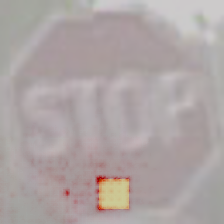}
\caption*{\footnotesize{Prototype 7}}
\end{minipage}%
\begin{minipage}{0.09\textwidth}
\includegraphics[scale=0.25,frame]{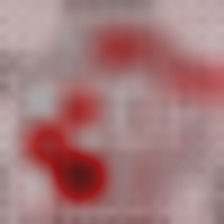}\\[1ex]
\includegraphics[scale=0.25,frame]{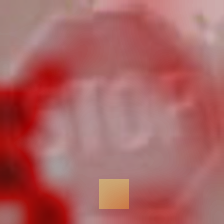}\\[1ex]
\includegraphics[scale=0.25,frame]{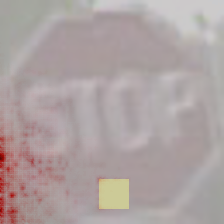}
\caption*{\footnotesize{Prototype 8}}
\end{minipage}%
\begin{minipage}{0.09\textwidth}
\includegraphics[scale=0.25,frame]{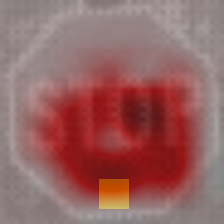}\\[1ex]
\includegraphics[scale=0.25,frame]{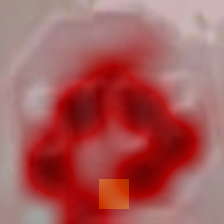}\\[1ex]
\includegraphics[scale=0.25,frame]{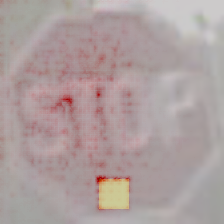}
\caption*{\footnotesize{Prototype 9}}
\end{minipage}%
\begin{minipage}{0.09\textwidth}
\includegraphics[scale=0.25,frame]{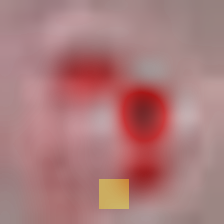}\\[1ex]
\includegraphics[scale=0.25,frame]{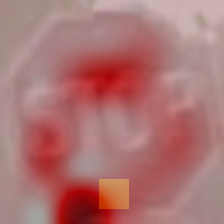}\\[1ex]
\includegraphics[scale=0.25,frame]{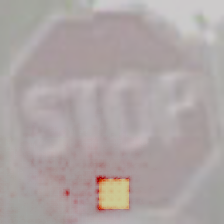}
\caption*{\footnotesize{Prototype 10}}
\end{minipage}%
\caption{\small{CH-50: Top row depicts the learned prototypes 1 to 10 for the stop sign class with Clever Hans in 50\% of the training data, the middle row depicts the ProtoPNet's heatmaps corresponding to the respective prototypes for the test image shown on the left, while the bottom row shows the corresponding PRP maps, which capture more precise information}.}
\label{fig:pt_stop50}
\end{figure*}

\begin{figure}[!htpb]
\centering
\begin{minipage}{0.12\textwidth}\centering
\includegraphics[scale=0.3]{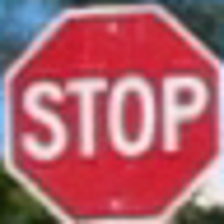}
\captionsetup{width=0.6\textwidth}
\caption*{\footnotesize{Test Image}}
\end{minipage}%
\begin{minipage}{0.12\textwidth}\centering
\includegraphics[scale=0.3,frame]{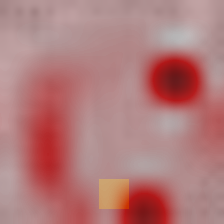}\\[1ex]
\includegraphics[scale=0.3,frame]{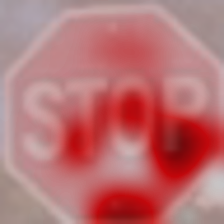}\\[1ex]
\includegraphics[scale=0.3,frame]{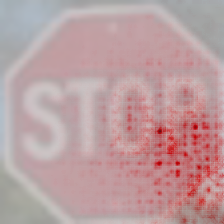}
\captionsetup{width=0.8\textwidth}
\caption*{\footnotesize{Prototype 4}}
\end{minipage}%
\begin{minipage}{0.12\textwidth}\centering
\includegraphics[scale=0.3,frame]{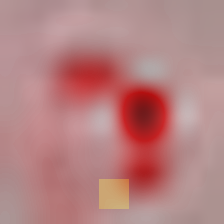}\\[1ex]
\includegraphics[scale=0.3,frame]{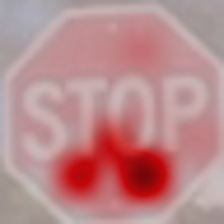}\\[1ex]
\includegraphics[scale=0.3,frame]{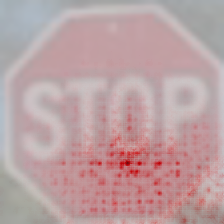}
\caption*{\footnotesize{Prototype 10}}
\end{minipage}%
\begin{minipage}{0.12\textwidth}\centering
\includegraphics[scale=0.3,frame]{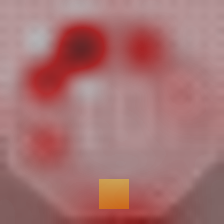}\\[1ex]
\includegraphics[scale=0.3,frame]{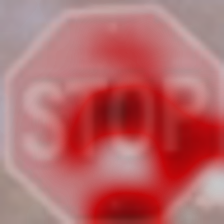}\\[1ex]
\includegraphics[scale=0.3,frame]{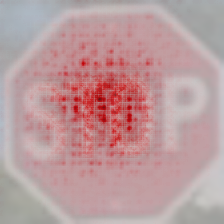}
\captionsetup{width=0.8\textwidth}
\caption*{\footnotesize{Prototype 2}}
\end{minipage}%
\\
\begin{minipage}{0.11\textwidth}
\footnotesize{Similarity Scores}
\end{minipage}
\begin{minipage}{0.12\textwidth}\centering
\footnotesize{6.672}
\end{minipage}
\begin{minipage}{0.12\textwidth}\centering
\footnotesize{5.773}
\end{minipage}
\begin{minipage}{0.12\textwidth}\centering
\footnotesize{4.184}
\end{minipage}
\caption{\small{CH-50: PRP Maps vs Activation Map Upsampling. 
The top three activated prototypes for the stop sign class for the test image are shown in the first row in descending order of similarity score. The second row shows the heatmaps generated by ProtoPNet and the last row shows the corresponding PRP maps.}}
\label{fig:LRPvsHM}
\end{figure}
In the following, we conduct an experiment, where we add a Clever Hans feature to the training dataset to investigate the difference between the heatmaps of ProtoPNet and the ones that PRP generates.
Therefore, we add the Clever Hans artifact to 50\% of the stop sign images in the training data (CH-50). The 10 prototypes for the stop sign class, learned by the ProtoPNet trained on the manipulated dataset, are shown in the first row of Figure \ref{fig:pt_stop50}.
Given a test image, shown at the very left of Figure \ref{fig:pt_stop50}, the heatmaps of ProtoPNet and the PRP heatmaps for the image are shown in the middle and bottom row of Figure \ref{fig:pt_stop50}. We can observe that the ProtoPNet heatmaps are coarse, highlighting wider areas in the test image, and that neighboring regions of the artifact are focused upon, rather than the precise location of the artifact.
In contrast, from the PRP maps, we can clearly observe that all prototypes are focusing precisely on the Clever Hans feature, some more (prototypes 2, 3, 4, 5, 7, 9, 10) and some less (prototypes 1, 6, 8). It is shown later that prototypes 6 and 8 are in fact not learning any significant features and even react strongly to random noise. With the new insight into the model behavior gained through the PRP maps, we can shed new light on the hypothesis from Section \ref{case_study_ch}. The idea was to remove the prototypes that had learned the Clever Hans, retrain the last layer and thus eliminate the Clever Hans effect. Given the original prototype explanation, this made sense, as only 2 of the 10 prototypes had learned the Clever Hans feature. With the PRP maps, however, we gain new knowledge and can see that all prototypes (some more, some less) take into account the Clever Hans feature, the yellow square.

We also note here that ProtoPNet heatmaps are highlighting all pixels in the image activated by different prototypes (before Max Pooling). If they were highlighting only the maximally activated region (after Max Pooling), it would only be able to depict connected regions in the image space, considering the naive upsampling heavily based on spatial location correspondence between the activation map and the input image. On the other hand, PRP maps represent the maximally activated pixels and are still able to highlight disjointed areas in the image, as can be seen in the PRP map for Prototype 5 in Figure \ref{fig:pt_stop50}, where both the artifact and ``ST" in the stop sign are shown as relevant. 

Figure \ref{fig:LRPvsHM} illustrates the difference between PRP maps and ProtoPNet heatmaps for a stop sign image with no artifact. PRP maps, as shown in the bottom row, are of higher resolution and, as noticed in this case, tend to show more accurate information than the normal upsampled heatmaps from ProtoPNet. PRP maps also contain higher variability, as shown by explanations for Prototype 2 and 4 in Figure \ref{fig:LRPvsHM}, which therefore yields more information from the original prototypes to explain the test pattern.

We now quantitatively evaluate the faithfulness of the PRP maps and ProtoPNet heatmaps regarding their ability to capture the most discriminative class-wise information. For this, we follow the strategy presented in \cite{ratedis}, referred to as the Relevance ordering test, where we start from a random image and monitor both the similarity and class scores as we gradually add the most relevant pixels to the image. 

\begin{figure}[!t]
\begin{subfigure}{.23\textwidth}
  \centering
  \includegraphics[width=1\linewidth]{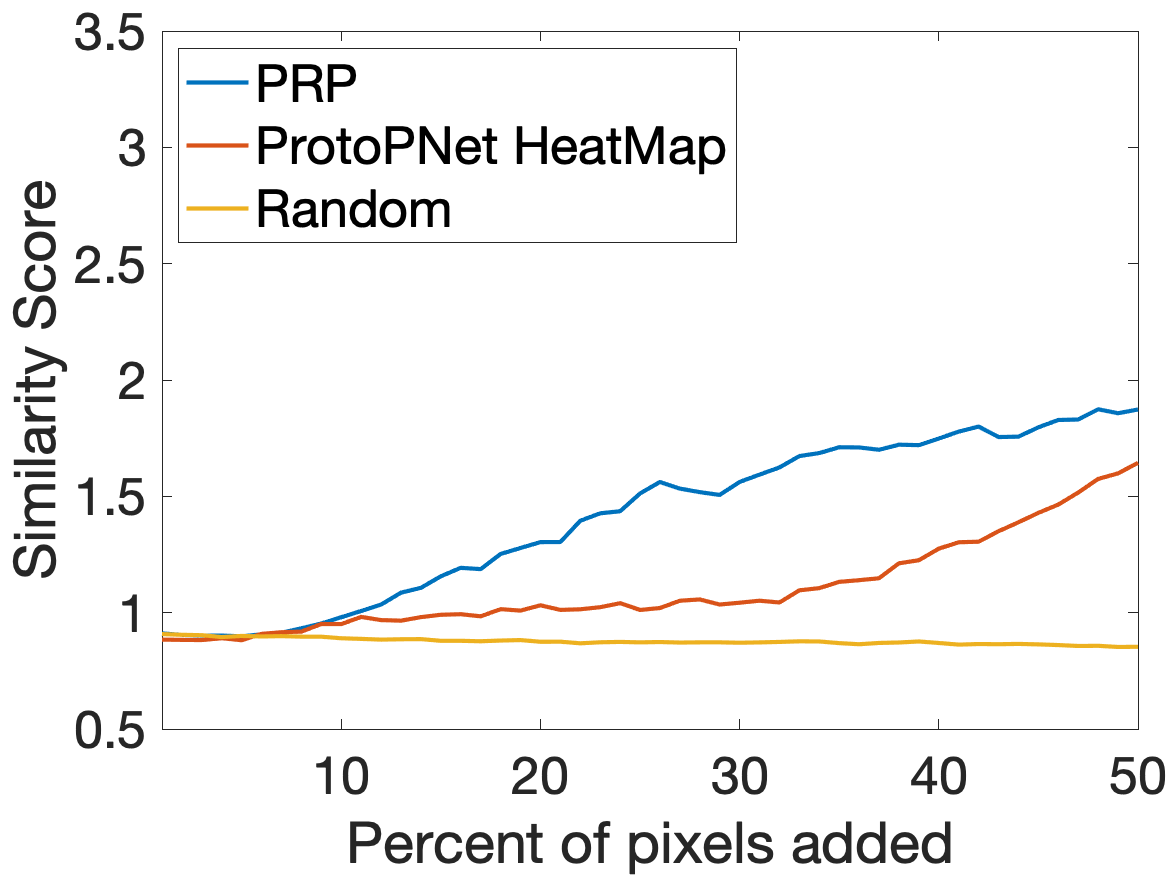}
  \caption{\footnotesize{Images without artifact}}
\end{subfigure}
\begin{subfigure}{.23\textwidth}
  \centering
  \includegraphics[width=1\linewidth]{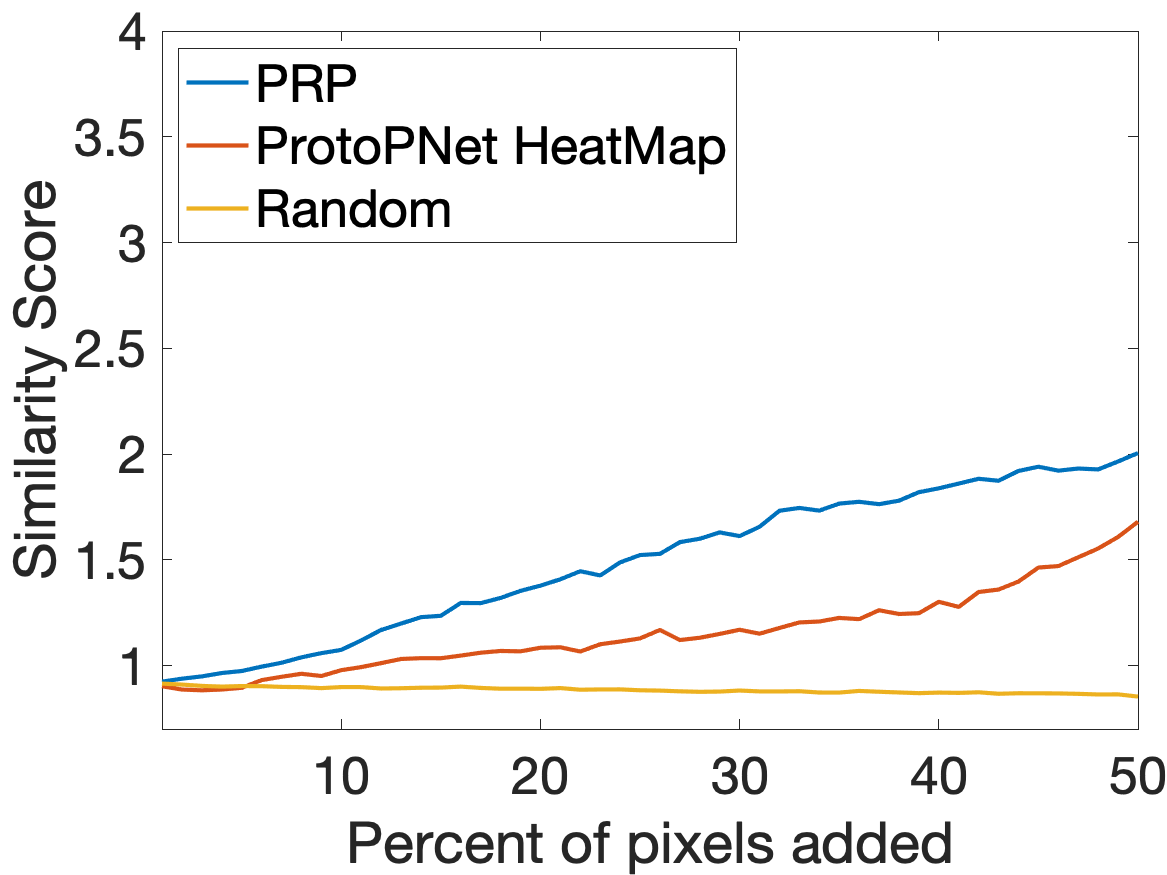}
  \caption{\footnotesize{Images with artifact}}
\end{subfigure} 
\caption{\small{CH-50: Quantitative evaluation of PRP Maps vs ProtoPNet Heatmaps via relevance ordering test. The results are shown as an average over all the prototypes and averaged over the same images without (left) and with artifact (right).}}
\label{fig:rd}
\end{figure}

\begin{figure}[!t]
\begin{subfigure}{.23\textwidth}
  \centering
  \includegraphics[width=1\linewidth]{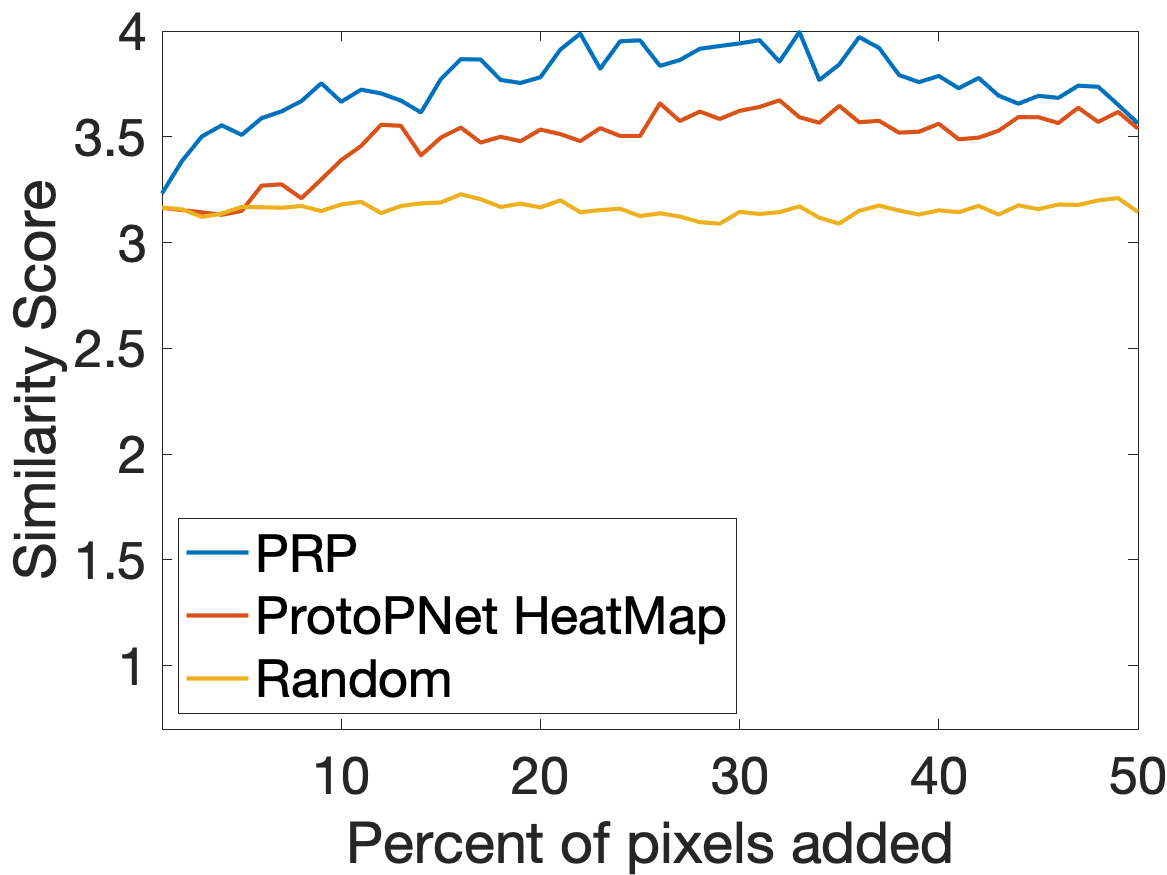}
  \caption{\footnotesize{Prototype 6}}
\end{subfigure}
\begin{subfigure}{.23\textwidth}
  \centering
  \includegraphics[width=1\linewidth]{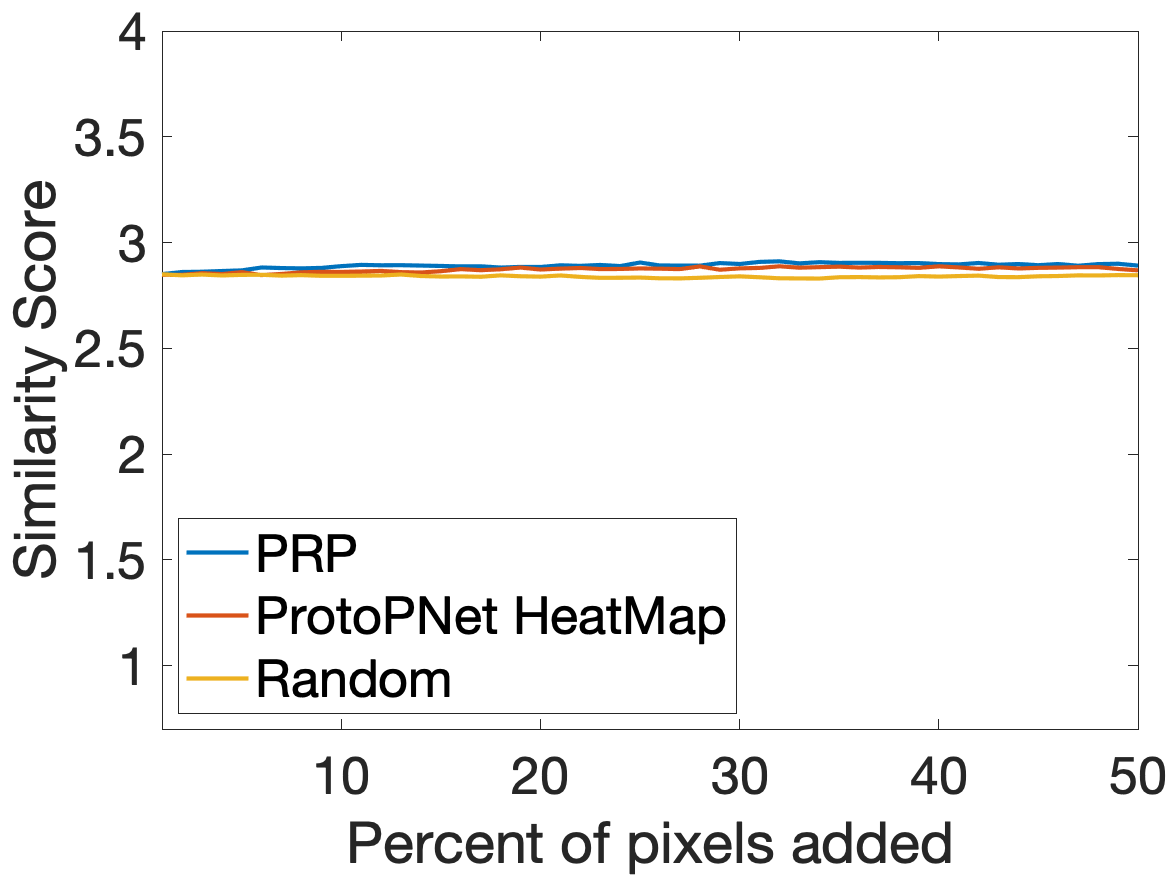}
  \caption{\footnotesize{Prototype 8}}
\end{subfigure} 
\caption{\small{CH-50: Relevance ordering test results shown for prototypes 6 and 8 of the stop sign class for the artifact test images. Both of these are not learning anything specific, therefore having high similarity with even random data.}}
\label{fig:rd_25_27}
\end{figure}

Primarily, we are interested in investigating if the most relevant pixels, according to the ProtoPNet heatmap and PRP map, activate the prototype the most. Hence, we are interested in measuring the similarity score between the activation map and prototype instead of the prediction value. For this, first, for an input image, PRP map and ProtoPNet heatmaps are computed, followed by sorting the pixels in descending order of their assigned relevance by PRP and ProtoPNet explanations, respectively. We then compute the similarity scores for different prototypes of the stop sign images while gradually adding the sorted pixels to a random image. We compute this for 50 randomly chosen clean images from the stop sign class and compute the average across all images followed by an average over all prototypes. The same experiment is repeated with the same images, this time adding the Clever Hans artifact. The average results for all prototypes of the stop sign class are shown in Figure \ref{fig:rd}. The x-axis represents the percentage of pixels that are replaced by the relevant pixels of the test image and the y-axis represents the corresponding similarity scores. As a baseline, we start from a random image and gradually replace a percentage of randomly chosen pixels by their test image pixel values and refer to this as the Random approach. From Figure \ref{fig:rd} we can observe that for both test case scenarios, i.e, the stop sign images with and without the artifact, PRP is able to capture more relevant information.
These quantitative results also uncover ineffective prototypes which are not learning anything specific from the training images and are reacting very highly even to random noise, as shown in Figure \ref{fig:rd_25_27}. This behavior is observed in both test scenarios of clean and artifact test data, with the results depicted for artifact test images in Figure \ref{fig:rd_25_27} for prototypes 6 and 8.

\begin{figure}[!t]
\begin{subfigure}{.23\textwidth}
  \centering
  \includegraphics[width=1\linewidth]{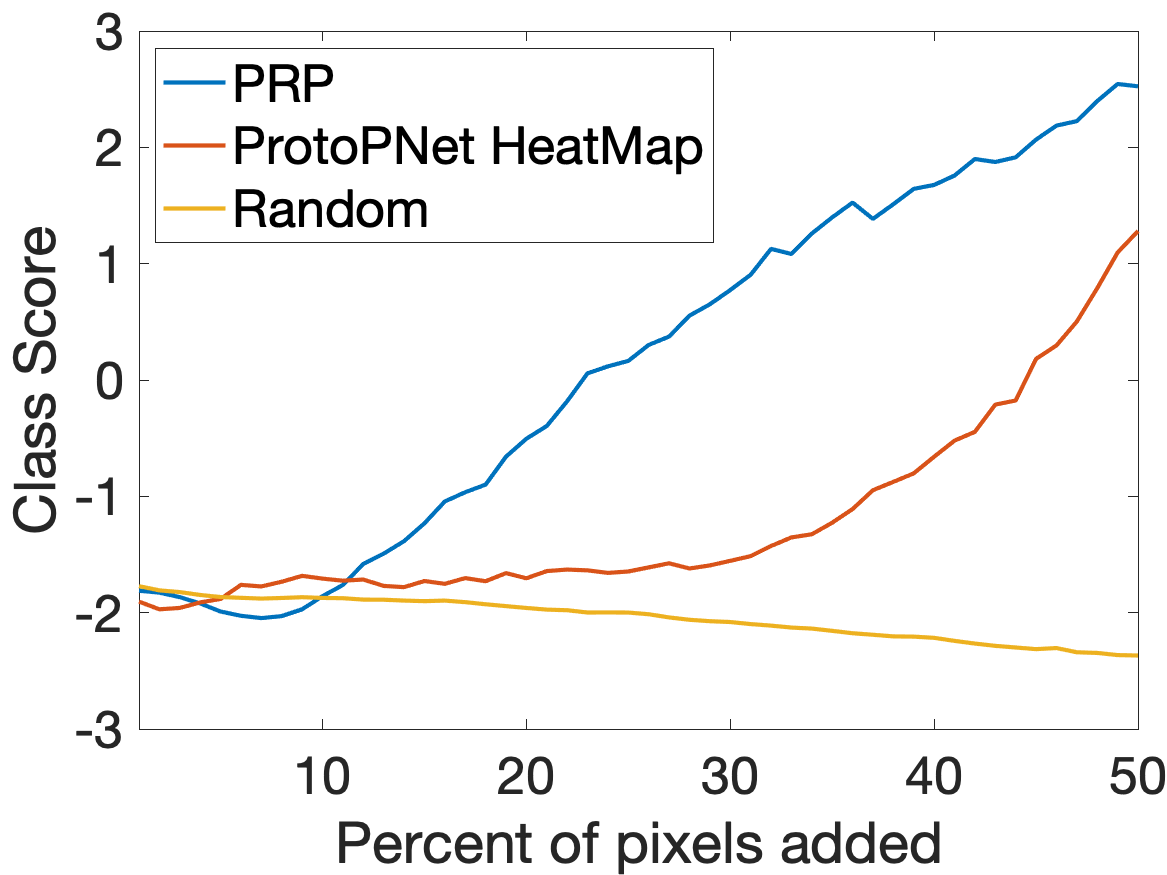}
  \caption{\footnotesize{Images without artifact}}
\end{subfigure}
\begin{subfigure}{.23\textwidth}
  \centering
  \includegraphics[width=1\linewidth]{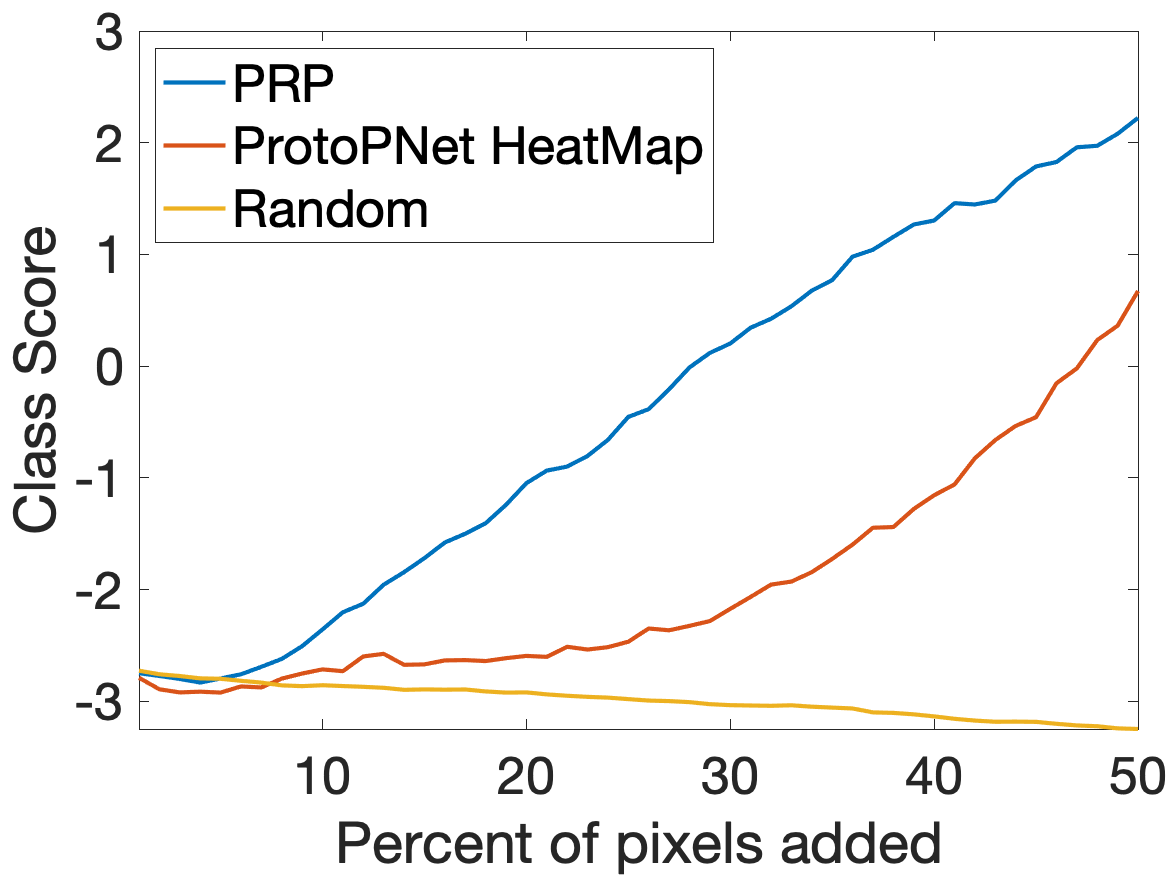}
  \caption{\footnotesize{Images with artifact}}
\end{subfigure}

\caption{\small{CH-50: Relevance ordering test for PRP vs ProtoPNet heatmaps based on the average class scores for the stop sign class prototypes averaged over the same images without (left) and with artifact (right).}}
\label{fig:rd_class}
\end{figure}
Until now, we quantitatively examined the faithfulness of PRP maps with respect to different prototypes in terms of the similarity score.
Additionally, we are interested in the faithfulness of the explanations when using the classification scores instead of the similarity scores.
Hence, for the same PRP and ProtoPNet explanation maps, we sort the pixels in descending order of relevance and gradually add them to the random images while monitoring the effect on the class score for the stop sign class as shown in Figure \ref{fig:rd_class}. 
We can observe from Figure \ref{fig:rd_class} that adding the most relevant pixels, based on the PRP explanations, results in a significantly steeper slope (orange line) than using the ProtoPNet heatmaps for both artifact (shown on the right) and clean dataset (shown on the left).
Therefore, conclusively, we can state that the relevance of the important discriminate features distributed by PRP is more accurate than by ProtoPNet explanations.
\begin{figure}[!t]
\centering
\begin{subfigure}{.45\textwidth}
\centering
    \includegraphics[scale=0.6]{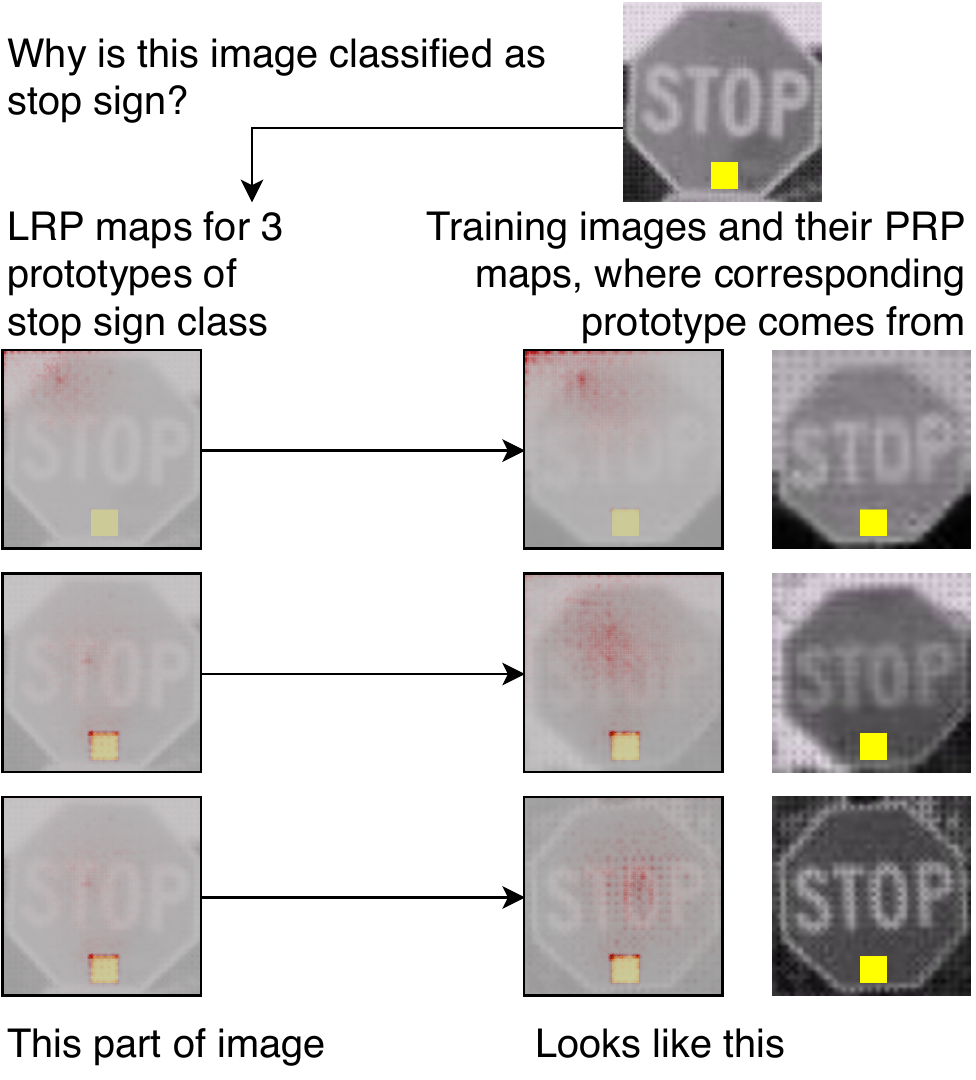}
\end{subfigure} \\[2ex]
\begin{subfigure}{.45\textwidth}
\centering
    \includegraphics[scale=0.6]{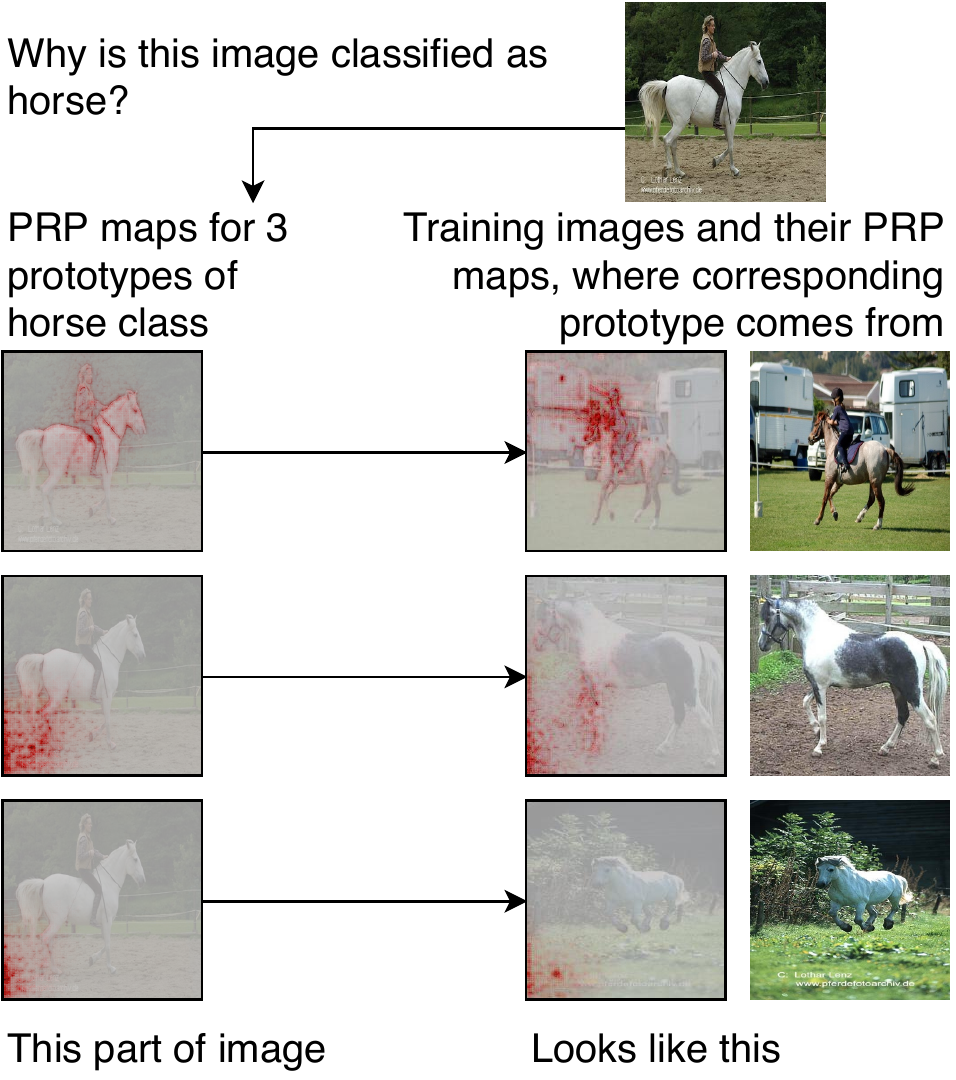}
\end{subfigure}
\caption{\small{\textit{This} looks \textit{more} like \textit{that}: Enhanced ProtoPNet self-explainability with PRP for a LISA stop sign image from the CH-50 dataset (top) and a PASCAL VOC horse image (bottom).}}
\label{fig:plrp}
\end{figure}
\subsection{Assessing the network behavior with PRP maps}
So far, we have established the drawbacks of ProtoPNet, which are the lack of higher resolution and spatially precise explanations which hinder the user in identifying the most relevant discriminative features. Accordingly, we proposed a method to overcome this lack of precise explanations. Our proposed PRP maps provide a higher level of fine grained explanations while keeping the benefit of ``this-looks-like-that" behavior of the ProtoPNet, as shown in Figure \ref{fig:plrp} for both LISA (CH-50) and PASCAL VOC 2007 datasets. Therefore, we still have inherent interpretability, where each class is being represented by a fixed number of prototypes. This exponentially reduces the need for the manual laborious task of analysing individual ad-hoc explainability heatmaps for assessing deep neural networks. Additionally, this also reduces the need to use semi-automated methodologies like SpRAy \cite{chans} to find patterns in a model's explanations with a huge number of explanation maps. 

\begin{figure}[!t]
    \centering
    \includegraphics[scale=0.47]{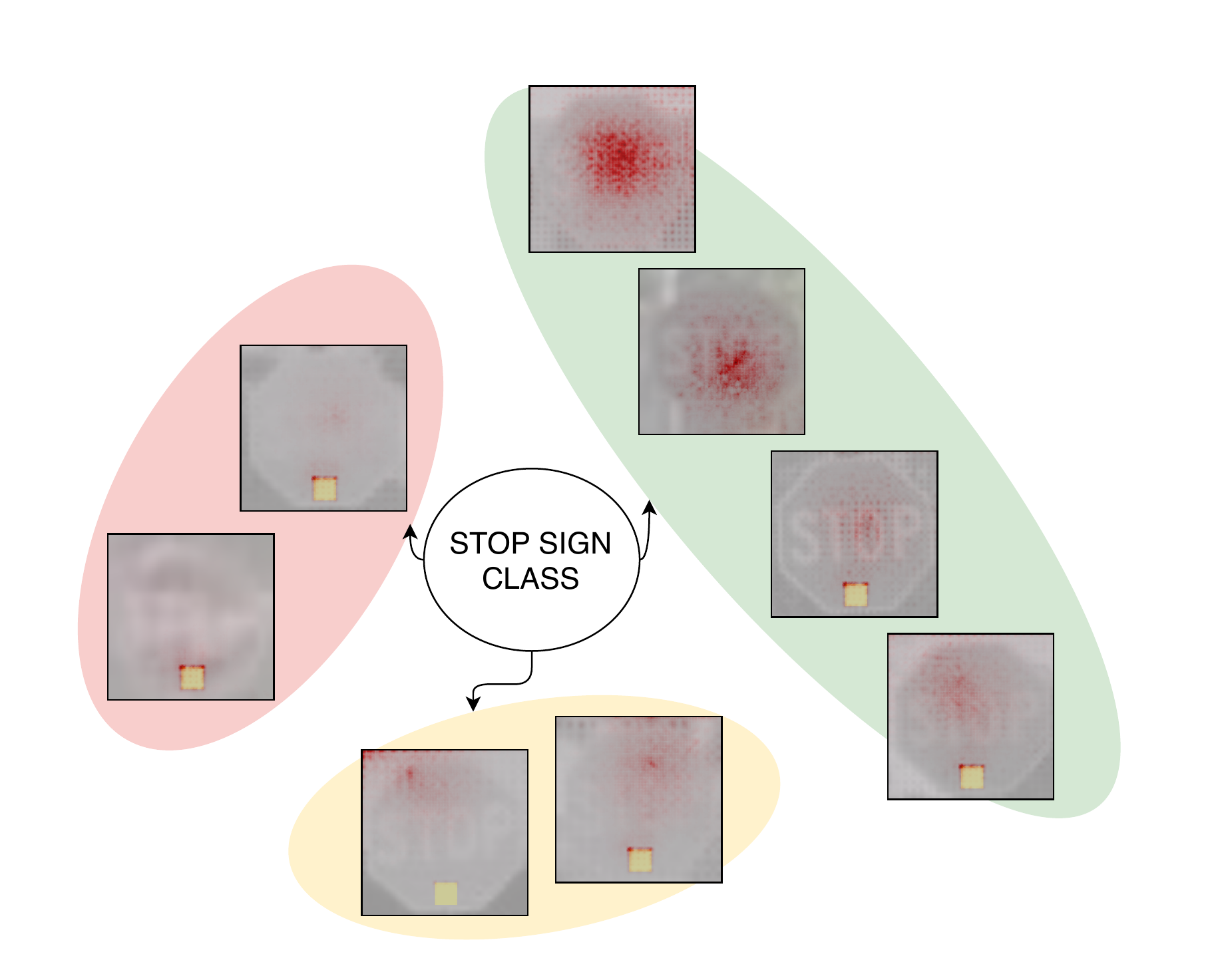}
    \caption{\small{CH-50: Representing prototypes for the stop sign class in clusters. The red cluster predominantly looks at the artifact, the green one looks at the text, while the yellow one looks at the corner features.}}
    \label{fig:ch50_cluster}
\end{figure}

We can now directly visually identify the strategies learned by the network by only looking at a few representative prototypes for each class. For instance, we manually cluster the PRP maps of the stop sign class for the LISA dataset, as shown in Figure \ref{fig:ch50_cluster}. We can observe, that aside from learning the artifact, the network is also relying on the textual part of the stop signs as well as on the corner features. Note, that we have excluded prototypes 6 and 8 from the assessment since they did not capture any useful information (see Figure \ref{fig:rd_25_27}).

\begin{figure}[!t]
\centering
\begin{minipage}{0.1\textwidth}\centering
\includegraphics[scale=0.28]{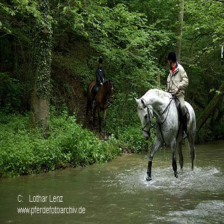}
\captionsetup{width=0.8\textwidth}
\caption*{\footnotesize{Test Image}}
\end{minipage}%
\begin{minipage}{0.1\textwidth}\centering
\includegraphics[scale=0.28,frame]{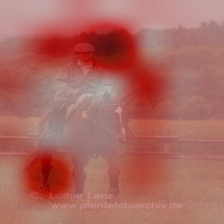}\\[1ex]
\includegraphics[scale=0.28,frame]{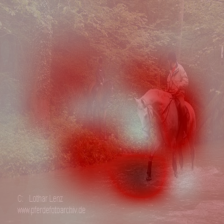}
\\[1ex]
\includegraphics[scale=0.28,frame]{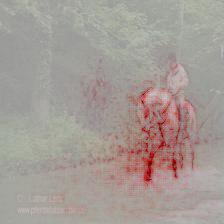}
\captionsetup{width=0.8\textwidth}
\caption*{\footnotesize{Prototype 6}}
\end{minipage}%
\begin{minipage}{0.1\textwidth}\centering
\includegraphics[scale=0.28,frame]{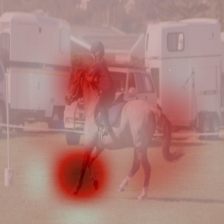}\\[1ex]
\includegraphics[scale=0.28,frame]{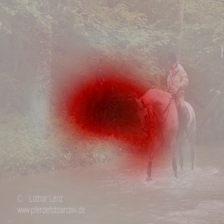}
\\[1ex]
\includegraphics[scale=0.28,frame]{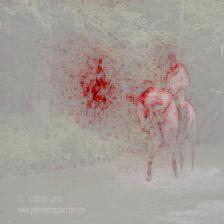}
\captionsetup{width=0.8\textwidth}
\caption*{\footnotesize{Prototype 4}}
\end{minipage}%
\begin{minipage}{0.1\textwidth}\centering
\includegraphics[scale=0.28,frame]{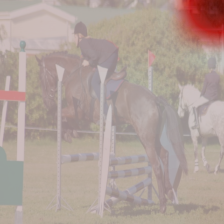}\\[1ex]
\includegraphics[scale=0.28,frame]{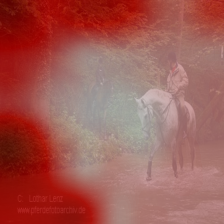}
\\[1ex]
\includegraphics[scale=0.28,frame]{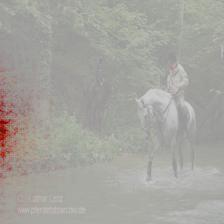}
\captionsetup{width=0.8\textwidth}
\caption*{\footnotesize{Prototype 2}}
\end{minipage}%
\begin{minipage}{0.1\textwidth}\centering
\includegraphics[scale=0.28,frame]{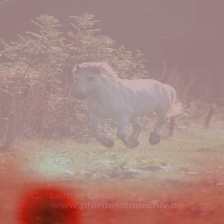}\\[1ex]
\includegraphics[scale=0.28,frame]{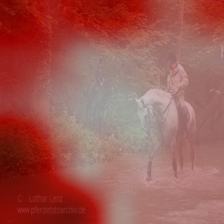}
\\[1ex]
\includegraphics[scale=0.28,frame]{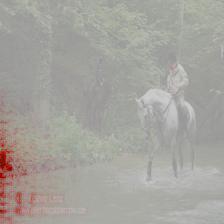}
\captionsetup{width=0.8\textwidth}
\caption*{\footnotesize{Prototype 3}}
\end{minipage}%
\\
\begin{minipage}{0.11\textwidth}
\footnotesize{Similarity Scores}
\end{minipage}
\begin{minipage}{0.085\textwidth}\centering
\footnotesize{5.509}
\end{minipage}
\begin{minipage}{0.085\textwidth}\centering
\footnotesize{1.971}
\end{minipage}
\begin{minipage}{0.085\textwidth}\centering
\footnotesize{0.999}
\end{minipage}
\begin{minipage}{0.085\textwidth}\centering
\footnotesize{0.962}
\end{minipage}
\caption{\small{PRP maps vs ProtoPNet heatmaps using the PASCAL VOC 2007 dataset: From top to bottom: Original prototypes, ProtoPNet Heatmaps and PRP maps for the top 4 activated horse class prototypes for the test image shown on the left. According to the PRP maps, prototype 6 is focusing at the horse features, prototype 4 is looking for the presence of a rider, prototype 2 is looking for grass in the background, and prototype 3 is looking at the Clever Hans watermark.}}
\label{pvoc_lrpvshm}
\end{figure}

\begin{figure}[!t]
    \centering
    \includegraphics[scale=0.47]{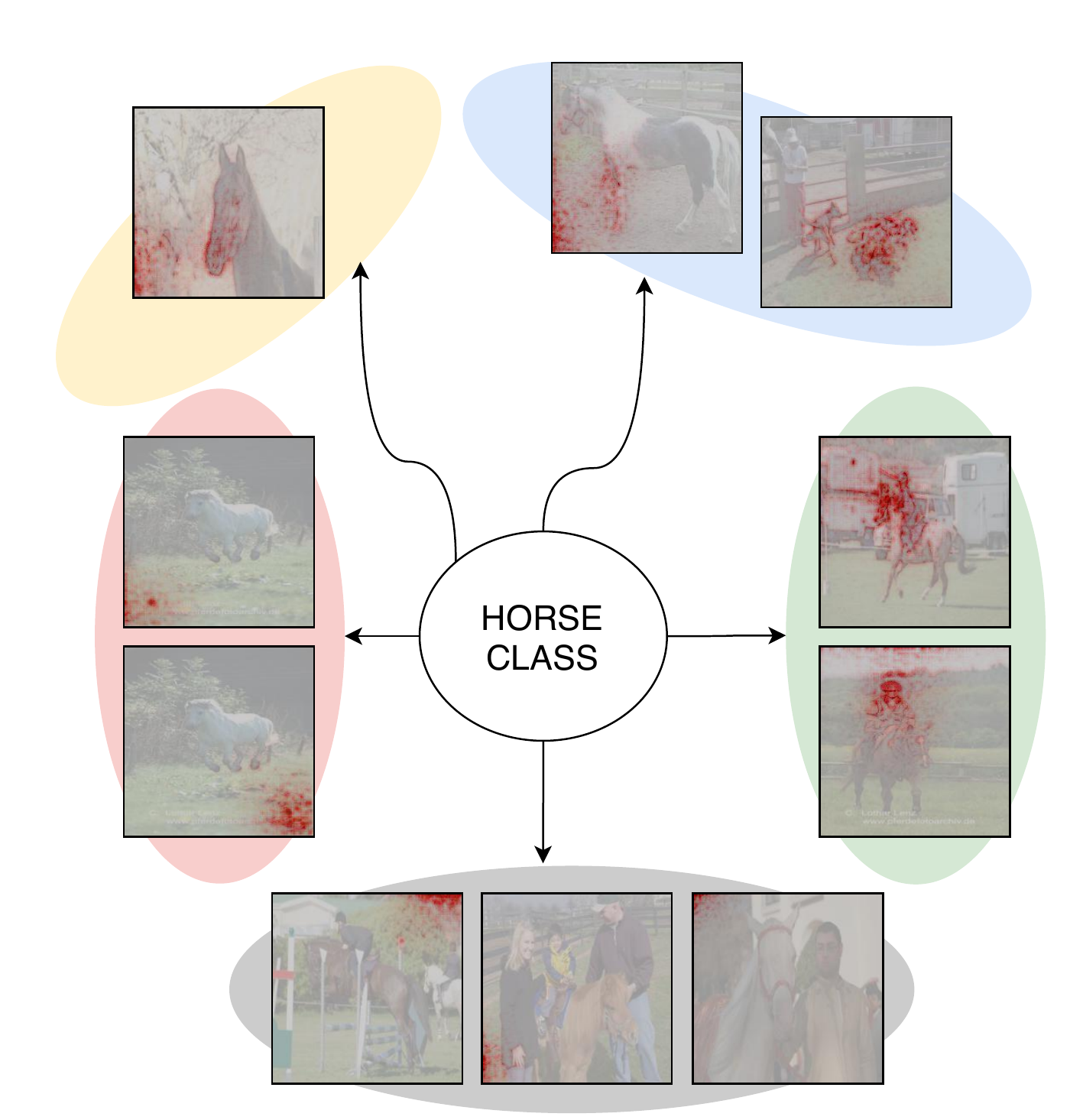}
    \caption{\small{Representing prototypes for the Horse class in clusters. Red clusters here looks at the "Clever-Hans" i.e, the watermark in the images, the yellow cluster looks at features of the horse's mouth, the blue one looks at the presence of horse-type legs, the green one looks if there is a rider present, and the gray one looks at the background features and is thus insignificant.}}
    \label{fig:PVOC_cluster}
\end{figure}

Following this, we investigate the performance of PRP and ProtoPNet explanations on the  PASCAL VOC 2007 dataset in order to uncover relevant features learned by the networks  for predicting the class horse. First, we show a few prototypes (top 4 activated) that were learned by the model for the horse class along with their ProtoPNet heatmaps and PRP Maps, shown in Figure \ref{pvoc_lrpvshm}. Here, we can observe that PRP explanations capture the relevant features in a more fine grained manner and are able to identify a Clever Hans strategy used by the model where it tends to focus on the text in the watermark in prototype 3, rather than on the horse. In contrast, the information in ProtoPNet's heatmaps in the second row of Figure \ref{pvoc_lrpvshm} is ambiguous since prototype 3 is allocating relevance to a broader background area. The strategies learnt by the network for recognizing a horse are grouped manually and visualized in Figure \ref{fig:PVOC_cluster}. The four effective groups, disregarding the insignificant gray cluster which focuses on the background features, represent the horse class in terms of a horse's face, legs, presence of a rider, and finally the Clever Hans watermark. 

\begin{table*}[!t]
\centering
\begin{tabular}{|c|c|c|c|c|c|c|c|c|c|c|c|c|c|c|}
\hline
&
\multicolumn{2}{c|}{\textbf{SpRAy-LRP}\cite{chans}}
 &
  \multicolumn{2}{c|}{\textbf{SpRAy-PRP}\cite{chans}} &
  \multicolumn{2}{c|}{\textbf{SiMVC}\cite{Trosten_2021_CVPR}} &
  \multicolumn{2}{c|}{\textbf{CoMVC}\cite{Trosten_2021_CVPR}} &
  \multicolumn{2}{c|}{\textbf{Co-Reg}\cite{coreg}} &
  \multicolumn{2}{c|}{\textbf{WMSC}\cite{wmvsc}} &
  \multicolumn{2}{c|}{\textbf{MCGC}\cite{mcgc}} \\ \cline{2-15} 
              & ACC & F1 & ACC     & F1   & ACC     & F1   & ACC            & F1         & ACC              & F1            & ACC              & F1            & ACC     & F1   \\ \hline
\textbf{CH-50} 
& 64.66\% & 0.45
& 53.59\% & 0.68 
& 99.43\% & 0.99 
& \textbf{100\%} & \textbf{1} 
& 99.57\%          & 0.99 
& 99.35\%          & 0.99  
& 67.53\% & 0.72
\\ \hline
\textbf{CH-20} 
& 72.91\% & 0.04
& 76.29\% & 0.08 
& \textbf{97.99}\% & \textbf{0.95} 
& 97.27\%        & 0.92      
& 94.54\%          & 0.86         
& 97.41\% & 0.93
& 80.03\% & 0.61 
\\ \hline
\textbf{BD-15} 
& 97.61\% & 0.91 
& 80.23\%  & 0.02
& 79.44\% & 0.60 
& 78.57\%        & 0.57      
& \textbf{99.42\%} & \textbf{0.98} 
& 99.35\%    & 0.98          
& 91.60\% & 0.76 
\\ \hline
\end{tabular}
\caption{\small{Accuracy (ACC) and F1-scores (F1) for different data scenarios with several multi-view clustering methodologies on PRP maps along with comparison with SpRAy on both PRP and LRP maps.}}
\label{tbl:accf1}
\end{table*}
\subsection{Multi-View Clustering for suppressing artifacts}
Artifacts in the data can be learned by the model, which subsequently might lead to the model exhibiting undesirable behavior, as shown in recent works \cite{chans} and demonstrated above in case of the self-explaining network, ProtoPNet. Thus, it is essential to either remove the artifacts from the data, or to ensure that the model is not basing its decisions on these spurious attributes present in the data.
We tried the latter in the experiments above via identifying and removing the artifact prototypes. However, as we observed, this is not possible since the artifact is not always perceivable by the ProtoPNet heatmaps even if the artifact was learned by a particular prototype. Using our suggested method, we are now able to find the prototypes that are activated by the artifact. It was further discovered using PRP in the previous sections, that almost all the prototypes incorporate the artifact features, thus suggesting the entanglement of the artifact information within the whole network. Therefore, instead of pruning the artifact prototypes, we propose to detect the samples in the training dataset that activate the artifact prototypes, which can be subsequently removed from the training data set before retraining the ProtoPNet on the cleansed dataset.

Using PRP, we obtain $k$ PRP maps corresponding to the artifact-containing class for each image, where $k$ corresponds to the number of learned prototypes for that class. We can consider these PRP maps as $k$ different views of the same image and can thus build on existing multi-view clustering methodologies to automatically cluster the training images and thereby discover clusters corresponding to artifact-containing images. In this work, we cluster the images into 2 clusters, an artifact and a clean data cluster.


To demonstrate the efficiency of PRP in detecting artifacts in the data, we test different multi-view clustering methodologies on the LISA dataset with 50\% and 20\% Clever Hans features added to the stop sign images. 
We further use the same methodologies for backdoor detection thereby demonstrating PRP's efficiency in multiple artifact scenarios. We also compare our clustering approach with SpRAy, which performs spectral clustering analysis on single view LRP maps, and demonstrate that our approach is able to capture better information in PRP maps, especially in the setting with multiple views. 

\subsubsection{Clever Hans type artifacts in 50\% training data}

The accuracy for CH-50 for the artifacts in the stop sign class in 100\% (artifact test) and 0\% (clean test) data is shown in Table \ref{tbl:ch_acc}. As can be seen, the accuracy for the stop sign class drops from 100\% to 94.6\% when there is no artifact in the test data. From Figure \ref{fig:pt_stop50}, prototypes 4 and 9 can be considered as ``artifact" prototypes according to ProtoPNet heatmaps. 
But as can be seen in Table \ref{tbl:ch_acc}, there is no effect on the artifact test accuracy when removing these prototypes. The same is true when removing the prototypes followed by retraining of the model. On the contrary, a decrease in the accuracy for the clean test data is observed. 
This again supports our assertion of misleading information provided by ProtoPNet's heatmaps. 

In order to obtain a clean data set, we aim to identify the samples that contain an artifact in the first place in order to remove them from the training set.
Assuming that the information on whether an artifact is present in a data point is recognizable in the PRP maps, we cluster the PRP maps in two clusters. For comparison, we use a set of representative algorithms to cluster the data, including SpRAy \cite{spray}, SiMVC \cite{Trosten_2021_CVPR}, CoMVC \cite{Trosten_2021_CVPR}, Co-Reg \cite{coreg}, WMSC \cite{wmvsc} and MCGC \cite{mcgc}.
We downsample the heatmaps to a size of 80x80, as this had negligible impact on the results and led to reduced computation time.

The overall accuracy of the clustering and F1-score for the artifact cluster are given in Table \ref{tbl:accf1}. We follow the experiments in \cite{Trosten_2021_CVPR} and train the SiMVC and CoMVC models for 100 epochs for 20 runs and report the results from the run resulting in the lowest unsupervised cost-function value.

As observed from Table \ref{tbl:accf1}, both SiMVC and CoMVC are working very efficiently to separate the artifact images from the clean images.
We also report the results for multi-view spectral clustering algorithms, i.e, Co-Reg, WMSC and MCGC in Table \ref{tbl:accf1}. Although being more computationally expensive, these algorithms 
are able to cluster the data effectively.
Co-Reg and WMSC always obtain an accuracy of above 94\% in separating the artifact data, and thus prove to be highly successful in detecting the artifacts. SiMVC and CoMVC on the other hand perform with almost 100\% accuracy when the artifact and non-artifact classes are balanced, i.e, in the current setting of CH-50.



\begin{table}[!b]
\centering
\begin{tabular}{rcc}
\multicolumn{3}{c}{Accuracy: 53.80\%}                                                   \\ \hline
\multicolumn{1}{|r|}{Cluster 1} & \multicolumn{1}{c|}{53} & \multicolumn{1}{c|}{643}   \\ \cline{2-3} 
\multicolumn{1}{|r|}{Cluster 2} & \multicolumn{1}{c|}{0}   & \multicolumn{1}{c|}{696} \\ \hline
\end{tabular}
\caption{\small{Best clustering obtained from PRP maps by prototype 7 of the stop sign class with CH-50 dataset. The F1-score for artifact cluster (Cluster 2) is 0.68.}}
    \label{tbl:spray_ch50}
\end{table}

\begin{figure*}[!htp]
\centering
\begin{minipage}{0.1\textwidth}\centering
\includegraphics[scale=0.25]{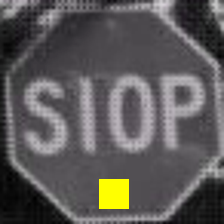}
\captionsetup{width=0.7\textwidth}
\caption*{\footnotesize{Test Image}}
\end{minipage}%
\begin{minipage}{0.09\textwidth}
\includegraphics[scale=0.25,frame]{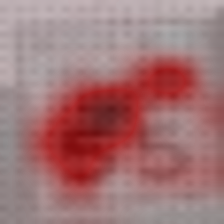}\\[1ex]
\includegraphics[scale=0.25,frame]{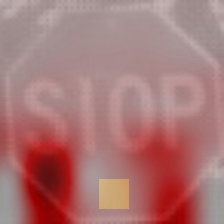}\\[1ex]
\includegraphics[scale=0.25,frame]{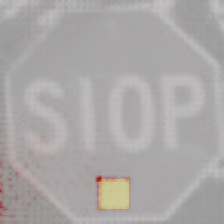}
\caption*{\footnotesize{Prototype 1}}
\end{minipage}%
\begin{minipage}{0.09\textwidth}
\includegraphics[scale=0.25,frame]{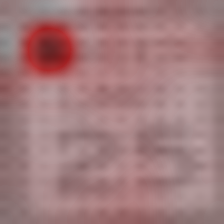}\\[1ex]
\includegraphics[scale=0.25,frame]{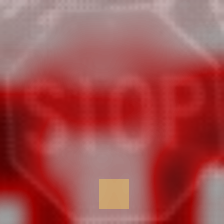}\\[1ex]
\includegraphics[scale=0.25,frame]{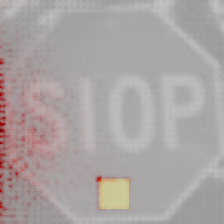}
\caption*{\footnotesize{Prototype 2}}
\end{minipage}%
\begin{minipage}{0.09\textwidth}
\includegraphics[scale=0.25,frame]{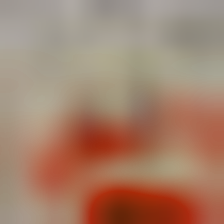}\\[1ex]
\includegraphics[scale=0.25,frame]{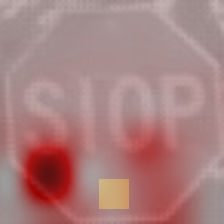}\\[1ex]
\includegraphics[scale=0.25,frame]{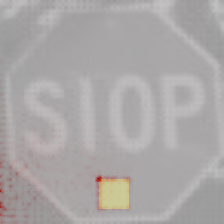}
\caption*{\footnotesize{Prototype 3}}
\end{minipage}%
\begin{minipage}{0.09\textwidth}
\includegraphics[scale=0.25,frame]{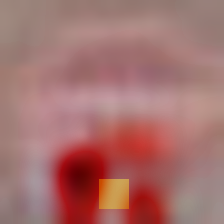}\\[1ex]
\includegraphics[scale=0.25,frame]{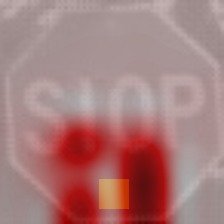}\\[1ex]
\includegraphics[scale=0.25,frame]{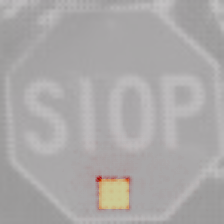}
\caption*{\footnotesize{Prototype 4}}
\end{minipage}%
\begin{minipage}{0.09\textwidth}
\includegraphics[scale=0.25,frame]{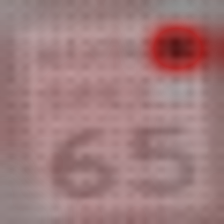}\\[1ex]
\includegraphics[scale=0.25,frame]{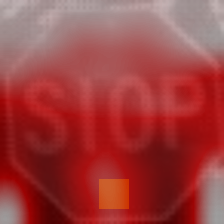}\\[1ex]
\includegraphics[scale=0.25,frame]{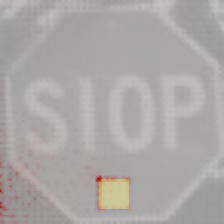}
\caption*{\footnotesize{Prototype 5}}
\end{minipage}%
\begin{minipage}{0.09\textwidth}
\includegraphics[scale=0.25,frame]{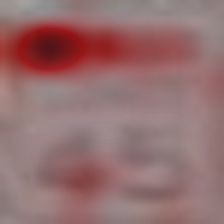}\\[1ex]
\includegraphics[scale=0.25,frame]{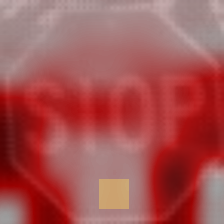}\\[1ex]
\includegraphics[scale=0.25,frame]{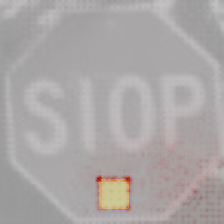}
\caption*{\footnotesize{Prototype 6}}
\end{minipage}%
\begin{minipage}{0.09\textwidth}
\includegraphics[scale=0.25,frame]{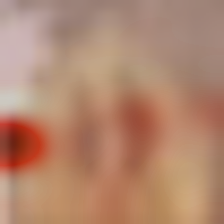}\\[1ex]
\includegraphics[scale=0.25,frame]{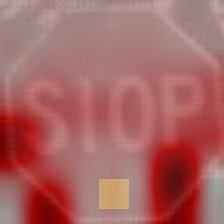}\\[1ex]
\includegraphics[scale=0.25,frame]{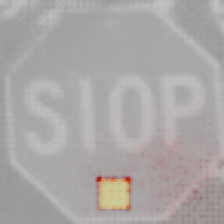}
\caption*{\footnotesize{Prototype 7}}
\end{minipage}%
\begin{minipage}{0.09\textwidth}
\includegraphics[scale=0.25,frame]{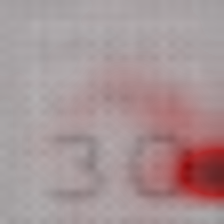}\\[1ex]
\includegraphics[scale=0.25,frame]{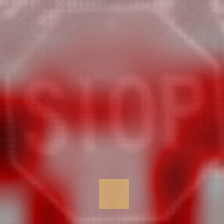}\\[1ex]
\includegraphics[scale=0.25,frame]{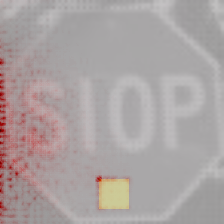}
\caption*{\footnotesize{Prototype 8}}
\end{minipage}%
\begin{minipage}{0.09\textwidth}
\includegraphics[scale=0.25,frame]{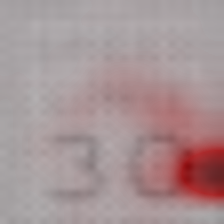}\\[1ex]
\includegraphics[scale=0.25,frame]{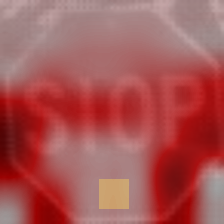}\\[1ex]
\includegraphics[scale=0.25,frame]{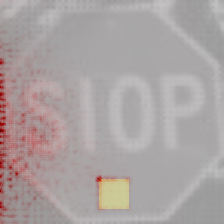}
\caption*{\footnotesize{Prototype 9}}
\end{minipage}%
\begin{minipage}{0.09\textwidth}
\includegraphics[scale=0.25,frame]{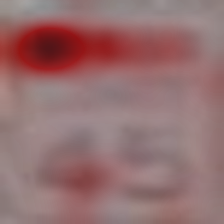}\\[1ex]
\includegraphics[scale=0.25,frame]{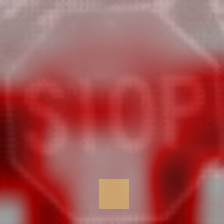}\\[1ex]
\includegraphics[scale=0.25,frame]{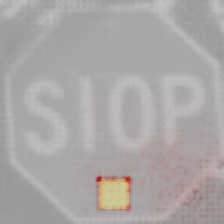}
\caption*{\footnotesize{Prototype 10}}
\end{minipage}%
\caption{\small{BD-15: Top row depicts the learned prototypes 1 to 10 for the speed limit class with the Backdoor in 15\% of the stop sign images (labeled as speed limit), the middle row depicts the ProtoPNet's heatmaps corresponding to the respective prototypes for the test image shown on the left and the bottom row shows the corresponding PRP maps for the prototypes, which capture more precise information.}}
\label{fig:pt_bd}
\end{figure*}

To compare against the multi-view clustering approaches,
we apply SpRAy \cite{chans}, on the LRP maps for the true class (SpRAy-LRP) as well as PRP maps for the prototypes of the true class (SpRAy-PRP). For SpRAy-LRP, we compute LRP maps using the rules in Section \ref{sec:prp}, followed by $\text{LRP}_\epsilon$ for the last layer and a combination of relevance for all prototypes. More details are provided in the Appendix \ref{appendix:spray-lrp}. Accordingly, we get one LRP map for each image, which is scaled down to 80x80 and flattened before applying SpRAy. For SpRAy-PRP, we combine the PRP map images by summing them across the channels and concatenating all 10 PRP maps for each image to get a 10x80x80 map. We then flatten it and apply SpRAy. 

The results for both are shown in Table \ref{tbl:accf1}. As observed, SpRAy fails in clustering the artifacts in CH-50 data using both LRP and concatenation of PRP maps. This behavior is expected since both SpRAy-LRP and SpRAy-PRP do not capture dependencies among multiple views of the same objects as opposed to other multi-view clustering methodologies. We also apply SpRAy on the individual prototype's PRP maps and report the best single-prototype performance in Table \ref{tbl:spray_ch50}, which is achieved by prototype 7. The obtained accuracy of 53.80\% and F1-score of 0.68, when compared to the combined SpRAy-PRP accuracy of 59.12\% and F1-score of 0.68, thus demonstrates that no individual prototype allows SpRAy to separate the artifact from the clean data.  

\subsubsection{Clever Hans type artifact in 20\% training data}
We also want to capture the scenarios when less Clever Hans artifacts are included in the training data. Therefore, we evaluate the efficiency of multi-view clustering methodologies on the unbalanced dataset CH-20.
The stop sign class accuracy for artifact and clean test data is 99.7\% and 95.8\%, respectively. This depicts that the stop sign class is still affected by the Clever Hans effect. 

Applying the multi-view clustering methodologies to this scenario, we report the accuracy and F1-score in Table \ref{tbl:accf1}. Results show that SiMVC is performing best with 97.99\% accuracy, with comparable performance by almost all the other multi-view clustering methods. SpRAy fails again with a very low F1-scores of 0.04 and 0.08 on LRP and PRP maps, thus suggesting clustering of almost all images into one cluster.



\subsubsection{Backdoor type artifact in 15\% training data}
Similar to the experiments above, we examine the backdoor setting, using the generated BD-15 dataset. The prototypes and their corresponding heatmaps for the speed limit class are shown in Figure \ref{fig:pt_bd}. The test accuracy when the artifact is present in 100\% of the stop sign test images is given in Table \ref{tbl:bd_acc}. Most of the stop sign images are now classified as speed limits and only 1\% of the stop sign images are classified correctly.

The prototypes of the speed limit class, as learned by ProtoPNet, show that only one prototype has learned the backdoor artifact, while all the remaining 9 prototypes correspond to the speed limit class, as shown in Figure \ref{fig:pt_bd}. As per ProtoPNet's explanations, removing prototype 4 of the speed limit class should solve the problem of backdoor attacks. We remove the prototype and retrain the last layer and report the accuracies in Table \ref{tbl:bd_acc}.

\begin{table}[!t]
\centering
\begin{tabular}{|c|ccc|}
\hline
 &
  \textbf{BD-15} &
  \textbf{\begin{tabular}[c]{@{}c@{}}BD-15\\ Remove \\ prototype 4\end{tabular}} &
  \textbf{\begin{tabular}[c]{@{}c@{}}BD-15\\ Retraining \\ last layer\end{tabular}} \\ \hline
\textbf{Artifact Test} &
  \multicolumn{1}{c|}{1.0\%} &
  \multicolumn{1}{c|}{6.5\%} &
  2.5\% \\ \hline
\textbf{Clean Test} &
  \multicolumn{1}{c|}{96.0\%} &
  \multicolumn{1}{c|}{96.0\%} &
  95.6\% \\ \hline
\end{tabular}
\caption{\small{Accuracy on the artifact test (backdoor in 100\% of the images in the stop sign class test data) and clean test data for BD-15, along with corresponding accuracies after removing the artifact prototype and retraining the last layer.}}
\label{tbl:bd_acc}
\end{table}

We can observe that removing the backdoor prototype has only a minor effect on the accuracy of the stop sign class, which increased from 1.0\% to 6.5\%. However, after retraining the last layer it again drops to only 2.5\%. This behaviour of the network thus emphasizes the inherent learning of the backdoor artifact by the network, which is not limited to only learning a specific backdoor prototype, as incorrectly suggested by ProtoPNet visualizations. Here, too, the PRP explanations decode the behavior of the model - they show that almost all prototypes are activated by the artifact, even if these prototypes refer to the speed limit signs. 

We therefore use multi-view clustering to clean the data of the backdoor feature and report the results in Table \ref{tbl:accf1}. 
SiMVC and CoMVC are still performing better than SpRAy-PRP with F1-scores of 0.60 and 0.57 respectively, as opposed to 0.02 F1-score of SpRAy-PRP. Although, SpRAy-LRP is performing good in this setting with a F1-score of 0.91. This is because LRP maps consist of negative relevances from the stop sign class in addition to the positive relevances from the speed limit class. This helps in accentuating the difference between speed limit and backdoor stop sign images.
Furthermore, all the multi-view spectral clustering-based algorithms are able to separate these clusters efficiently, with the best being Co-Reg with an accuracy of 99.42\% and a F1-score of 0.98.

\section{Conclusion}
Considering the success of machine learning algorithms in diverse safety-critical applications, it is instrumental to verify the behavior of these models. 
In this work, we assess the faithfulness of the explanations provided by a well known self-explainable network, ProtoPNet, and provide an in-depth assessment of its behavior in the presence of a range of artifacts. Our results indicate that, despite the attractiveness of self-explaining models, they are still very far from achieving the required quality of explanations. 
Considering this, we propose a model-aware method, PRP, to generate more precise and higher resolution prototypical explanations.
These enhanced explanations help in uncovering more credible decision strategies, while keeping the self-explainability intact. We further show that these explanations are able to uncover the spurious artifact features learned by the model, which are then efficiently identified and removed via our proposed multi-view clustering strategy.
The insights obtained in this work highlight the importance of evaluating the quality of self-explaining machine learning approaches and will pave the way towards the development of more robust and precise models, thereby increasing their trustworthiness.

{\small
\bibliographystyle{unsrt}
\bibliography{ref}
}

\section{Appendix}
\subsection{ProtoPNet: Cost function}
\label{appendix:ppnet}
The overall cost function for ProtoPNet is:
\begin{align}
\mathcal{L}_{\text{total}} = \mathcal{L}_\text{CE} + \lambda_{\text{clst}} \mathcal{L}_\text{clst} + \lambda_\text{sep} \mathcal{L}_\text{sep} 
\end{align}
$\mathcal{L}_\text{CE}$ is the cross entropy (CrsEnt) loss, $\mathcal{L}_\text{clst}$ is the cluster loss and $\mathcal{L}_\text{sep}$ is the separation loss, defined as:
\begin{align}
    \mathcal{L}_\text{CE} &= \min_{{W}} \frac{1}{N} \sum_{i=1}^{N} \text{CrsEnt}(\mathbf{\hat{y}}_i,\mathbf{y}_i) \\
    \mathcal{L}_\text{clst} &= \frac{1}{N} \sum_{i=1}^{N} \min_{m:\mathbf{p}_m\in \mathbf{P}_{\mathbf{y}_i}} \min_{\mathbf{\widetilde{z}}} ||\mathbf{\widetilde{z}} - \mathbf{p}_m ||^2_2 \\
    \mathcal{L}_\text{sep} &= - \frac{1}{N} \sum_{i=1}^{N} \min_{m:\mathbf{p}_m\notin \mathbf{P}_{\mathbf{y}_i}} \min_{\mathbf{\widetilde{z}}} ||\mathbf{\widetilde{z}} - \mathbf{p}_m ||^2_2 
\end{align}
where $N$ are the total number of training images, $\mathbf{y}_i$ is the true label for image $i$, $\mathbf{\hat{y}}_i$ is the predicted label, $W$ represents the learnable parameters of the whole network, $\mathbf{P}_{\mathbf{y}_i}$ are all the prototypes belonging to class $\mathbf{y}_i$ and $\mathbf{\widetilde{z}}$ are the patches of the convolutional output which are of the same size as the prototypes.

\subsection{SpRAy-LRP}
\label{appendix:spray-lrp}
For SpRAy based on LRP maps, we first backpropagate the output relevances i.e, class scores to the similarity score layer. We follow the $\textbf{LRP}_{CMP}$ rule and use the $\text{LRP}_\epsilon$ rule \cite{towards}:
\begin{equation}
    \mathbf{R}_{i\xleftarrow{}j}^{(l,l+1)} = \frac{z_{ij}}{z_j + \epsilon \: . \:
     \text{sign}(z_j)
    }
    \mathbf{R}_{j}^{(l+1)}
\end{equation}
For the rest of the network, the rules for PRP are used. Considering that we are now computing relevance corresponding to all the prototypes, we combine them to get the relevance at $CONV$ layer as:
\begin{equation}
    \mathbf{R}^{(CONV,AM)}_{ijc} = \sum_{m=1}^{n} \mathbf{R}^{(CONV,AM)}_{mijc}
\end{equation}

\subsection{LISA 5 class dataset}
\label{appendix:lisa5}
\begin{table}[H]
\centering
\begin{tabular}{|p{1.3cm}|l|}
\hline
\textbf{Restriction signs} & \begin{tabular}[c]{@{}l@{}}noRightTurn, keepRight, thruMergeLeft, thruMergeRight,\\ thruTrafficMergeLeft, doNotPass, noLeftTurn, \\ doNotEnter, rightLaneMustTurn\end{tabular}                                                                                                                                                                                                                \\ \hline
\textbf{Speed limits}      & \begin{tabular}[c]{@{}l@{}}speedLimit40, speedLimit25, speedLimit35, speedLimit50, \\ speedLimit45, truckSpeedLimit55, speedLimit65, speedLimit55, \\ speedLimit30, speedLimit15, schoolSpeedLimit25\end{tabular}                                                                                                                                                                              \\ \hline
\textbf{Stop signs}        & \begin{tabular}[c]{@{}l@{}}stopAhead,  stop\end{tabular}                                                                                                                                                                                                                                                                                                                                             \\ \hline
\textbf{Warning signs}     & \begin{tabular}[c]{@{}l@{}}turnLeft, signalAhead, zoneAhead25, school, curveLeft, \\ pedestrianCrossing, curveRight, rampSpeedAdvisory50, \\ rampSpeedAdvisoryUrdbl, dip, rampSpeedAdvisory40, \\ merge, turnRight, slow, roundabout, speedLimitUrdbl, \\ zoneAhead45, intersection, laneEnds, rampSpeedAdvisory45, \\ rampSpeedAdvisory20, rampSpeedAdvisory35, addedLane\end{tabular} \\ \hline
\textbf{Yield signs}        & \begin{tabular}[c]{@{}l@{}}yield, yieldAhead\end{tabular}                                                                                                                                                                                                                                                                                                                                           \\ \hline
\end{tabular}
\caption{\small{Combination of classes from LISA dataset for 5-class CH-100, CH-50, CH-20 and BD-15 datasets.}}
\end{table}

\end{document}